\documentclass[final,3p,times]{elsarticle}
\usepackage{amssymb}
\usepackage{array}
\usepackage{booktabs}
\usepackage{makecell}
\usepackage{xltabular}
\keepXColumns

\newcolumntype{C}[1]{>{\centering\arraybackslash}p{#1}}
\newcolumntype{Y}{>{\raggedright\arraybackslash}X}
\newcolumntype{Z}{>{\centering\arraybackslash}X}

\newcommand{\TopHead}[1]{%
  \parbox[t]{\linewidth}{\centering\strut #1\strut}%
}
\usepackage{algorithm}
\usepackage{algpseudocode}

\usepackage{algorithm}
\usepackage{comment}
\usepackage{color}
\usepackage{color,soul}
\usepackage{enumitem}
\setlist[itemize]{itemsep=2pt, topsep=2pt, parsep=2pt}
\setlist[enumerate]{itemsep=2pt, topsep=2pt, parsep=2pt}
\usepackage{makecell}
\usepackage{mathtools}
\usepackage{multirow}
\usepackage{nomencl}
\usepackage{threeparttable}
\usepackage{bbm}
\usepackage{subfigure}

\usepackage{etoolbox}
\usepackage{xurl}
\usepackage[
    colorlinks=true,
    linkcolor=blue,
    urlcolor=black,
    citecolor=green,
    unicode=true,
    bookmarksnumbered=true,
    bookmarks=true,
    bookmarksdepth=3
]{hyperref}
\newif\ifrevision
\revisionfalse   % false

\usepackage{xcolor}
\usepackage[normalem]{ulem} % for \sout, underline without breaking \emph
\usepackage{xargs}          % optional, for multiple optional args

\usepackage{textcomp}

\usepackage{booktabs}
\journal{Applied Energy}

\begin{document}

\begin{frontmatter}

  %% Title, authors and addresses

  \title{A Dual-Positive Monotone Parameterization for Multi-Segment Bids and a Validity Assessment Framework for Reinforcement Learning Agent-based Simulation of Electricity Markets}

  \author[label1]{Zunnan Xu}
  \author[label1]{Zhaoxia Jing\corref{cor1}}
  \cortext[cor1]{Corresponding author.}
  \ead{zxjing@scut.edu.cn}
  \author[label2]{Zhanhua Pan}

  \address[label1]{School of Electric Power Engineering, South China University of Technology, Guangzhou, 510640, Guangdong, China}
  \address[label2]{Department of Engineering, University of Exeter, Exeter EX4 4PY, U.K.}

  \begin{abstract}
    Reinforcement learning agent-based simulation (RL-ABS) has become an important tool for electricity market mechanism analysis and evaluation. In the modeling of monotone, bounded, multi-segment stepwise bids, existing methods typically let the policy network first output an unconstrained action and then convert it into a feasible bid curve satisfying monotonicity and boundedness through post-processing mappings such as sorting, clipping, or projection. However, such post-processing mappings often fail to satisfy continuous differentiability, injectivity, and invertibility at boundaries or kinks, thereby causing gradient distortion and leading to spurious convergence in simulation results. Meanwhile, most existing studies conduct mechanism analysis and evaluation mainly on the basis of training-curve convergence, without rigorously assessing the distance between the simulation outcomes and Nash equilibrium, which severely undermines the credibility of the results. To address these issues, this paper proposes: (i) three necessary conditions, NC1–NC3, that post-processing mappings in RL-ABS should satisfy, in order to characterize what kinds of post-processing mappings will not distort the gradient signals associated with the actually executed actions; (ii) the Dual-Positive Monotone Parameterization (DPMP) method, which first lets the policy network output two positive vectors corresponding to generation output and price increments, and then constructs a continuously differentiable, injective, and invertible mapping between the policy-network outputs and the feasible set of stepwise bid curves through normalized cumulative summation and cumulative summation of nonnegative increments, thereby avoiding distortion in gradient propagation; and (iii) a two-level Validity Assessment Framework for RL-ABS, in which the first level evaluates validity at the single-agent algorithm level by taking the theoretical optimal profit as the benchmark and constructing the optimality gap indicator, while the second level introduces the exploitability metric at the multi-agent simulation level to assess the distance between simulation outcomes and Nash equilibrium by freezing the opponents’ policies and training an approximate best response. Experimental results show that, in the single-agent setting, DPMP reduces the steady-state relative optimality gap to 3.26\% ± 0.73\%, significantly outperforming the baselines based on sorting, clipping, and projection. Moreover, DPMP can be consistently integrated with several mainstream algorithms, including A2C, TRPO, PPO, and DDPG. In the IEEE 39-bus network-constrained multi-agent simulation scenario, the multi-agent strategy profile based on DPMP-PPO achieves a maximum exploitability of 1.266\% and an average of approximately 0.20\%, exhibiting steady-state characteristics close to an ε-Nash equilibrium. Overall, the DPMP and Validity Assessment Framework proposed in this paper can provide substantially stronger support for conclusions drawn from RL-ABS-based electricity market mechanism analysis, thereby offering more reliable decision support for future electricity market mechanism design and evaluation.
  \end{abstract}


\end{frontmatter}

\section{Introduction}
\label{sec:1}

In electricity markets, generators submit multi-segment bids composed of price--generation pairs. Such bids must satisfy monotonicity and price-cap requirements, and the associated bidding strategies directly affect market-clearing outcomes and market equilibrium \cite{song2025optimal}. Therefore, an accurate representation of this bid structure is essential for market mechanism design and for evaluating how market rules shape participant behavior \cite{glismann2021ancillary,ringler2016agent}.Reinforcement learning agent-based simulation (RL-ABS) has been widely used for this purpose. It can capture bounded rationality \cite{pan2023multi}, agent heterogeneity, and dynamic learning processes \cite{sridhar2024residential}. It also complements optimization-based models \cite{ventosa2005electricity,baillo2004optimal} and game-theoretic equilibrium models \cite{hobbs2000strategic} in electricity market analysis and evaluation \cite{shafie2014stochastic,fraunholz2021advanced}. However, most existing RL-ABS studies still rely on simplified bid models, such as single-segment bids or scalar markup coefficients imposed on marginal cost curves \cite{nanduri2007reinforcement}. These approaches cannot represent the full price--generation bid structure required in real-world markets. As a result, they reduce the fidelity of electricity market simulation and weaken the validity of the conclusions drawn from such studies.

\subsection{Deficiencies of Bid Models in Current RL-ABS}
\label{sec:1.1}

\begin{figure}[H]
    \centering
    \includegraphics[width=\linewidth]{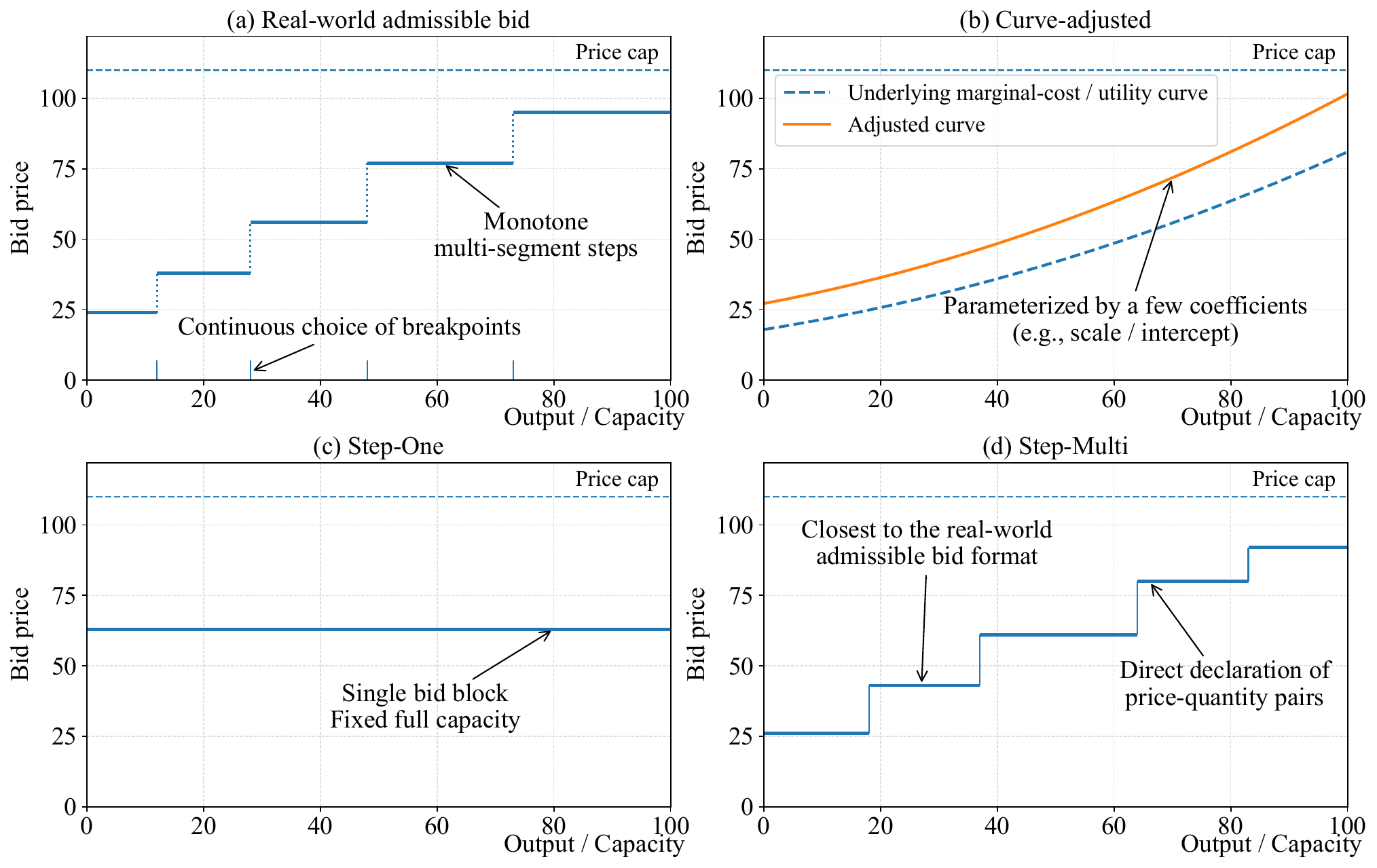}
    \caption{Conceptual sketches of real-world electricity-market bids and common RL-ABS bid models.}
    \label{fig:bid_model}
\end{figure}

As shown in Fig.~\ref{fig:bid_model}(a), bid models in real-world electricity markets should satisfy three requirements \cite{song2025optimal}: (i) bids consist of multiple price--generation pairs and must be monotonically non-decreasing, thereby forming a stepwise bid curve; (ii) bid prices are bounded by a price floor and a price cap; (iii) both bid prices and generation output are defined on continuous domains. However, as summarized in Table~\ref{tab:agent_bid_models}, existing reinforcement learning agent-based simulation (RL-ABS) studies did not adopt this bid structure from the outset. Instead, the field evolved gradually from low-dimensional simplified representations to higher-dimensional continuous ones. Accordingly, the bid models used by agents can be broadly classified into three categories, as illustrated in Fig.~\ref{fig:bid_model}(b)(c)(d).

The first category is Curve-adjusted (Fig.~\ref{fig:bid_model}(b)). In this category, agents do not directly determine the multi-segment price–generation pairs of a bid. Instead, they adjust a marginal cost curve, a utility function, or a predefined bid curve using a small number of discrete or continuous coefficients, such as an overall markup coefficient or an intercept term, and the adjusted curve is then submitted as the bid. Earlier studies in this category mainly relied on tabular reinforcement learning. Rahimiyan et al.~\cite{rahimiyan2010adaptive} proposed a Q-learning-based bidding method with market power identification and fuzzy adaptive parameter tuning to improve the long-term profits of generators in dynamic electricity markets. Yu et al.~\cite{yu2023reinforcement} proposed a Hysteretic Q-learning method for settings with incomplete information and renewable-energy uncertainty, with the aim of improving market-clearing efficiency and approaching Nash equilibrium. Later studies adopted deep reinforcement learning to address more complex market environments. Liang et al.~\cite{liang2020agent} developed a DDPG-based reinforcement learning agent-based simulation (RL-ABS) approach to more stably approximate Nash equilibrium and characterize collusion under incomplete information. Chandrakala et al.~\cite{chandrakala2023multi} proposed a multi-agent cooperative bidding method based on Variant Roth--Erev learning to improve generator profits and congestion-pricing decisions in the day-ahead market. Rokhforoz et al.~\cite{rokhforoz2023multi} introduced a multi-agent DDPG bidding method integrated with GCNs to improve generator profits and enhance generalization to changes in network topology. Yin et al.~\cite{yin2024multi} proposed a PER-enhanced multi-agent continuous bidding method to simulate a centralized bilateral-auction electricity market and improve bidding performance in complex and uncertain environments. Weng et al.~\cite{weng2025optimizing} combined GCNs and DDPG in a multi-agent bidding framework to model correlations among market participants and improve strategic bidding profits in bilateral electricity markets. Despite differences in algorithms and application scenarios, these studies share one common feature: the agent adjusts the bid curve as a whole through a few coefficients, rather than directly determining each segment price and generation output independently. Although such methods are convenient for modeling and training, their bidding expressiveness remains much weaker than that of the monotone multi-segment bids required in real-world electricity markets.

The second category is Step-One (Fig.~\ref{fig:bid_model}(c)). In this category, the agent submits only a single segment price, while the generation output is fixed at the unit’s maximum capacity or at a scenario-specified constant. The price action space can be either discrete or continuous. Wu et al.~\cite{wu2024intelligent} developed a MAS-DRL bidding framework and compared several value-based methods, including DQN, DDQN, and Async n-step Q-learning, under a single-segment pricing setting with fixed generation output. Zhang et al.~\cite{zhang2024game} adopted a continuous single-segment bid model to study game-theoretic bidding and profit allocation for VPP participation in joint energy and ancillary service markets. Although these studies differ in whether the price action space is discrete or continuous, they share the same basic limitation: only the price dimension remains decision-dependent, whereas generation output is fixed ex ante. As a result, Step-One models cannot represent the full joint decision structure of price and generation-output breakpoints required in real-world electricity markets.

The third category is Step-Multi (Fig.~\ref{fig:bid_model}(d)). In this category, agents submit multi-segment price--generation pairs. The action spaces of both prices and generation output can be either discrete or continuous, and the submitted prices must satisfy monotonicity constraints and the price cap. Jiang et al.~\cite{jiang2024optimal} studied the multi-period, multi-segment bidding strategy of a price-maker VPP in the day-ahead market, and characterized market equilibrium and profit optimization through multi-agent reinforcement learning. Pan et al.~\cite{pan2025decision} attempted to develop a GenCo bidding and cost-modeling method that more closely reflects real-world market rules, with the aim of supporting future electricity market mechanism design. Song et al.~\cite{song2025optimal} proposed an optimal bidding framework for high-dimensional real-time markets, reducing the computational difficulty of high-dimensional bidding problems while improving profits. Compared with the first two categories, these studies are clearly closer to the bid format used in real-world electricity markets, because they incorporate monotonicity constraints and the price cap.

Meanwhile, the type of action space in these three bid models is closely related to the choice of algorithmic implementation. Specifically, when the action space is discrete, model-based or value-based algorithms are typically adopted; when the action space is continuous, policy-based methods are more commonly employed\cite{sutton1998reinforcement}\cite{zhao2025mathematical}.

\begin{figure}[H]
    \centering
    \includegraphics[width=\linewidth]{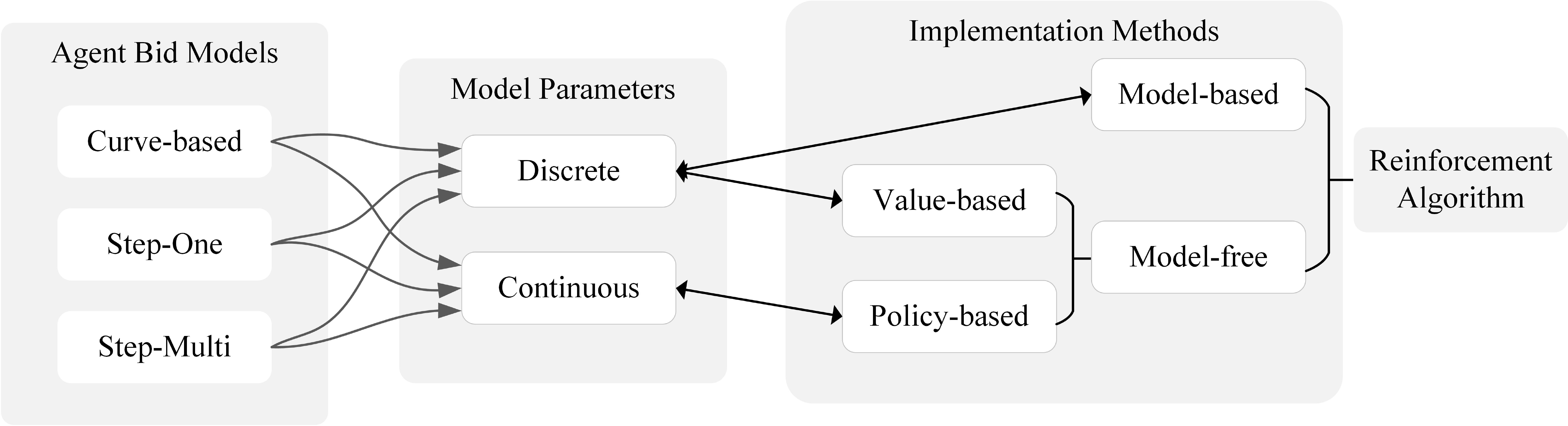}
    \caption{Relationship between Agent Bid Models and Reinforcement Learning Methods.}
    \label{fig:Relationship between Agent Bid Models and Reinforcement Learning Methods}
\end{figure}

{
\scriptsize
\setlength{\tabcolsep}{3.0pt}
\renewcommand{\arraystretch}{1.10}

\begin{xltabular}{\textwidth}{
C{0.82cm}
C{0.78cm}
C{2.40cm}
C{1.75cm}
C{3.45cm}
C{2.65cm}
Z}
\caption{Agent bid models of previous studies.}
\label{tab:agent_bid_models}\\

\toprule
\TopHead{Ref.} &
\TopHead{Year} &
\TopHead{Agent bid model} &
\TopHead{Parameter\\type} &
\TopHead{Whether satisfying\\real-world market\\bid requirements} &
\TopHead{Algorithm} &
\TopHead{Post-processing\\mapping\\(for Step-Multi\\continuous type)} \\
\midrule
\endfirsthead

\multicolumn{7}{c}{\tablename\ \thetable\ (continued)}\\
\toprule
\TopHead{Ref.} &
\TopHead{Year} &
\TopHead{Agent bid\\model} &
\TopHead{Parameter\\type} &
\TopHead{Whether satisfying\\real-world market\\bid requirements} &
\TopHead{Algorithm} &
\TopHead{Post-processing\\mapping\\(for Step-Multi\\continuous type)} \\
\midrule
\endhead

\bottomrule
\endfoot

\bottomrule
\endlastfoot

\cite{song2025optimal} & 2025 & Step-Multi & Continuous & Satisfying all & DDPG &
projection\\
\addlinespace[3pt]

\cite{pan2025decision} & 2025 & Step-Multi & Continuous & \makecell[c]{Not satisfying\\Constraint iv} &
\makecell[c]{DRL\\(Actor--Critic\\framework)} &
clipping \\
\addlinespace[3pt]

\cite{weng2025optimizing} & 2025 & Curve-adjusted & Continuous & \makecell[c]{None satisfied} &
DDPG-GCN &
-- \\
\addlinespace[3pt]

\cite{jiang2024optimal} & 2024 & Step-Multi & Continuous & \makecell[c]{Not satisfying\\Constraint iv} &
\makecell[c]{TD3\\(compared with\\DDPG and PPO)} &
clipping \\
\addlinespace[3pt]

\cite{yin2024multi} & 2024 & Curve-adjusted & Continuous & \makecell[c]{None satisfied} &
DDPG-PER &
-- \\
\addlinespace[3pt]

\cite{wu2024intelligent} & 2024 & Step-One & Discrete & \makecell[c]{None satisfied} &
\makecell[c]{DQN / DDQN /\\Async n-step\\Q-learning} &
-- \\
\addlinespace[3pt]

\cite{zhang2024game} & 2023 & Step-One & Continuous & \makecell[c]{None satisfied} &
DDPG &
-- \\
\addlinespace[3pt]

\cite{rokhforoz2023multi} & 2023 & Curve-adjusted & Continuous & \makecell[c]{None satisfied} &
DDPG-GCN &
-- \\
\addlinespace[3pt]

\cite{yu2023reinforcement} & 2023 & Curve-adjusted & Discrete & \makecell[c]{None satisfied} &
\makecell[c]{Hysteretic\\Q-learning} &
-- \\
\addlinespace[3pt]

\cite{chandrakala2023multi} & 2023 & Curve-adjusted & Discrete & \makecell[c]{None satisfied} &
\makecell[c]{Variant\\Roth--Erev\\(VRE)} &
-- \\
\addlinespace[3pt]

\cite{liang2020agent} & 2020 & Curve-adjusted & Continuous & \makecell[c]{None satisfied} &
DDPG &
-- \\
\addlinespace[3pt]

\cite{rahimiyan2010adaptive} & 2010 & Curve-adjusted & Discrete & \makecell[c]{None satisfied} &
\makecell[c]{Adaptive\\Q-learning} &
-- \\

\end{xltabular}

}

Given that real-world electricity market bidding rules follow the Step-Multi Continuous bid model \cite{manual202111}, policy gradient methods are the most suitable choice for this setting, because they can directly handle multidimensional continuous decisions. However, to generate bids that satisfy the requirements of multi-segment structure, monotonicity constraints, boundedness constraints, and continuous decision-making, the raw output of the policy network is usually transformed into a feasible bid through post-processing mappings such as sorting \cite{lohndorf2013optimizing}, clipping \cite{pan2025decision}, or projection \cite{song2025optimal,jiang2024optimal}. Specifically, sorting rearranges the segment prices in nondecreasing order to enforce monotonicity constraints. Clipping truncates or recursively adjusts segment prices that violate boundedness or monotonicity constraints so that they fall back into the admissible range. Projection maps the entire raw output vector, under a specified distance criterion, onto the feasible set that satisfies both monotonicity and boundedness constraints. Although these methods are convenient in engineering implementation, they may destroy the continuously differentiable, injective, and invertible relationship between policy outputs and executed actions. As a result, the policy gradient may no longer correspond strictly to the actually executed action, thereby causing gradient distortion, objective mismatch, and spurious convergence.

In fact, this issue is not unique to RL-ABS. Similar problems also arise in other constrained-action settings, such as robotic continuous control and safe reinforcement learning, where post-processing is used mainly to enforce simple boundary constraints or safety constraints \cite{fujita2018clipped}. However, electricity market bidding is more challenging because multi-segment bids must satisfy monotonicity constraints, boundedness constraints, and multidimensional structural constraints at the same time. Therefore, the inconsistency between the executed action and the action implied by the gradient is more complex and more pronounced in electricity market settings. This is precisely the problem that the present work aims to address.

\subsection{Insufficient validity verification in current RL-ABS research}

For electricity market mechanism analysis and evaluation, final conclusions often rest on a tacit premise: the strategic interactions generated by RL-ABS converge to a Nash equilibrium. Only under this premise can differences in simulation outputs be interpreted as differences in equilibrium outcomes induced by different market mechanisms \cite{yu2023finding}. This issue has received much broader attention in the literature on game learning and multi-agent reinforcement learning. For example, OpenSpiel, a general game-learning framework developed by DeepMind \cite{lanctot2019openspiel}, uses tools such as exploitability and best response to assess whether a strategy profile is close to equilibrium.

However, most existing RL-ABS studies in electricity markets focus on market mechanism design, the convergence of agent profit curves, or comparisons among algorithms. Much less attention has been paid to a more fundamental issue: whether the simulation results are sufficiently valid to support mechanism comparison. In particular, there is still a lack of validity assessment for practical applications of reinforcement learning to electricity market simulation, especially quantitative measures of the distance between simulation outcomes and Nash equilibrium. As a result, directly using RL-ABS results to compare alternative market mechanisms may lead to conclusions with limited credibility.

Moreover, a few recent studies have begun to examine Nash equilibrium more explicitly in electricity market simulation, but their scope remains narrow. For example, \cite{yu2023finding} analyzed whether strategies can converge to Nash equilibrium by combining reinforcement learning with a distributed clearing algorithm under a single-parameter supply-function framework. \cite{yu2023reinforcement} incorporated incomplete information and renewable uncertainty into the bidding game, and examined whether Bayesian Nash equilibrium can be reached by combining Bayesian games with reinforcement learning. Overall, however, these studies either rely on highly simplified low-dimensional bid parameterizations, restrictive type assumptions, and specialized convergence-analysis frameworks, or remain focused on equilibrium reachability within specific theoretical models. They therefore do not answer the key question in more general and practically relevant RL-ABS settings: how far simulation outcomes remain from Nash equilibrium.

\subsection{Main contributions of this work}

To address two key limitations of existing reinforcement learning agent-based simulation (RL-ABS)—continuous multi-segment bid representation and validity assessment of simulation results—and to improve its fidelity, verifiability, and the credibility of conclusions in electricity market mechanism analysis and evaluation, this paper makes the following main contributions:

(a) Motivated by the implementation requirements of monotone multi-segment bids in electricity markets, we formally derive the necessary conditions that post-processing mappings should satisfy under policy gradient methods, and show how common operations such as sorting, clipping, and projection can induce gradient distortion and learning bias.

(b) We propose a Dual-Positive Monotone Parameterization (DPMP) method that enables joint continuous decision-making over segment prices and generation output. While satisfying monotonicity, boundedness, and feasibility constraints, DPMP preserves a continuously differentiable, injective, and invertible mapping from policy outputs to the feasible bid space.

(c) We develop a two-level Validity Assessment Framework for RL-ABS. At the single-agent level, the optimality gap is used to evaluate how close the learned strategy is to the theoretical optimum. At the multi-agent level, exploitability is used to measure how far a strategy profile deviates from an $\varepsilon$-Nash equilibrium.

(d) We conduct systematic experiments in an IEEE 39-bus network-constrained market. The experiments cover different post-processing mappings, policy gradient methods, and multi-agent game settings. The results verify the applicability of the proposed method to electricity market mechanism research and support the credibility of the results.

Overall, this paper contributes not only a new bid parameterization method, but also a quantitative framework for assessing equilibrium validity. Together, these contributions improve the interpretability of RL-ABS and the credibility of the results in electricity market mechanism research. They also provide a more reliable basis for future electricity market design, rule comparison, and policy evaluation.

\section{Research Problem and Research Framework}

\subsection{The multi-segment bidding decision problem in electricity markets}

In studies of electricity market mechanisms, the core behavior of a generation entity is to submit bid curves that satisfy both physical and institutional constraints under prescribed market rules. In day-ahead markets, or more generally in electricity energy markets characterized by centralized clearing, generation entities are typically required to submit stepwise bids composed of multiple price levels and their corresponding generation-output segments. Such bids must not only satisfy basic requirements such as the price cap and unit capacity constraints, but also often need to preserve an economically meaningful monotone structure, thereby reflecting the variation of unit marginal costs with output as well as the restrictions imposed by market bidding rules on admissible bid forms.  

In electricity market simulation, if the multi-segment bids used in real-world markets are simplified into an overly coarse bidding format, the strategy space that market participants can express will be compressed, and the information on which market clearing is based will consequently be distorted. As a result, the locational marginal prices, generator profits, and even equilibrium outcomes obtained from the simulation may deviate from those that would arise under the real-world bidding rules, thereby biasing subsequent conclusions regarding market rules or differences across market mechanisms\cite{song2025optimal}.

Compared with simplified bid forms such as Curve-adjusted or Step-One, continuous multi-segment stepwise bids are closer to the bidding format used by generators in real-world electricity markets, because they preserve decision freedom in both segment prices and generation output breakpoints. This richer representation offers greater expressive power, but also entails substantially higher modeling complexity:

\begin{equation}
\begin{aligned}
a_i &= (\mathbf{q}_i,\mathbf{p}_i)\in \mathcal{A}_i,\\
\mathcal{A}_i
&= \left\{(\mathbf{q}_i,\mathbf{p}_i)\,\middle|\, 
0 \le q_{i,1} \le \cdots \le q_{i,K} \le \bar{Q}_i,\;
0 \le p_{i,1} \le \cdots \le p_{i,K} \le \bar{P}
\right\}.
\end{aligned}
\label{eq:action_space}
\end{equation}

Here, $\mathbf{q}_i=(q_{i,1},\cdots,q_{i,K})$ denotes the vector of generation output breakpoints for unit $i$, and $\mathbf{p}_i=(p_{i,1},\cdots,p_{i,K})$ denotes the corresponding vector of segment prices. $\bar{Q}_i$ and $\bar{P}$ represent the unit capacity limit and the market price cap, respectively. It is therefore evident that continuous multi-segment bids are, in essence, a high-dimensional continuous decision problem jointly constrained by capacity limits, price-cap limits, and monotonicity constraints.

\subsection{Modeling challenges in learning constrained continuous bids}

The modeling of continuous multi-segment bids first raises a method-selection problem. For this type of joint price-output decision task, a discrete-action formulation typically requires combinatorial discretization across multiple bidding segments, which not only causes the action space to expand rapidly, but also weakens the model’s ability to represent genuinely continuous bidding behavior. By contrast, policy gradient methods can characterize policy distributions directly in a continuous action space and are therefore better suited to the joint continuous decision problem involved in multi-segment bidding. Hence, in the Step-Multi: Continuous setting that more closely matches real-world bidding rules, adopting a policy-gradient paradigm is not merely an empirical preference; rather, it is a modeling choice that follows naturally from the continuity and structural properties of the problem itself.

However, adopting policy gradient methods also brings important challenges. The raw outputs generated by the policy network do not, in general, naturally satisfy monotonicity constraints, boundedness constraints, or feasibility constraints. Existing studies therefore commonly rely on post-processing mappings, such as sorting, clipping, or projection, to transform raw actions into executable bids.

\begin{equation}
\begin{aligned}
x_t &\sim \mu_\theta(\cdot \mid s_t),\\
a_t &= h(x_t)\in \mathcal{A},\\
J(\theta) &= \mathbb{E}_{\tau}\!\left[\sum_{t=0}^{T-1}\gamma^t r(s_t,a_t)\right].
\end{aligned}
\label{eq:policy_mapping}
\end{equation}

Here, $x_t$ denotes the raw continuous output generated by the policy network under state $s_t$, and $h(\cdot)$ denotes the post-processing mapping that transforms the raw output into the feasible bid space $\mathcal{A}$. The environmental reward and state transition are ultimately determined by the executed action $a_t$, rather than by the raw output $x_t$. Although such a treatment is convenient in engineering implementation, it introduces a more fundamental modeling inconsistency: the object being optimized by the policy is the network’s raw output, whereas the object to which the environment actually responds is the post-processed feasible bid. If $h(\cdot)$ does not possess satisfactory continuity, injectivity, or local invertibility, then the gradient signals received by the policy parameters may no longer accurately correspond to the actual executed actions, thereby leading to gradient distortion, objective mismatch, and spurious convergence.

Against this backdrop, this paper first investigates the necessary conditions that a post-processing mapping should satisfy under the policy gradient framework, with the aim of answering the following question: in constrained continuous bid learning, what kind of action representation can avoid distorting the actual executed actions on which learning should be based? Building on this analysis, we propose DPMP, which enables policy outputs to be mapped into the feasible bid space in a continuously differentiable and injective manner. In this way, the inconsistency between the learning object and the executed object can be eliminated at its source.

\subsection{Validity concerns in RL-ABS-based electricity market simulation}

Even if the representation problem of constrained continuous bids is resolved, the task of RL-ABS in electricity market research remains incomplete. The reason is that the purpose of electricity market simulation is not merely to obtain a high-profit agent, but to use simulation outcomes to analyze market rules, compare alternative market mechanisms, and discuss their equilibrium implications. In this context, improvements in profit curves or stability during training can at most indicate that an algorithm has achieved some form of numerical convergence under a given workflow; they do not automatically imply that the simulation results already possess the interpretability required for mechanism analysis.

In the existing literature, one long-standing yet insufficiently emphasized issue is that simulation results are often directly used for mechanism comparison once training appears to have converged, while little systematic examination is conducted as to whether they still remain far from the theoretically rational state. If the learned outcomes still deviate substantially from the theoretical optimum, or if the multi-agent policy profile has not yet approached a game-theoretic equilibrium, then the observed differences across market mechanisms may in fact be confounded by algorithmic error, defects in action representation, or inadequate training, rather than being attributable solely to the market rules themselves. In other words, trainability does not imply suitability for mechanism analysis, and profit convergence does not imply credibility of the results.

To address this issue, this paper further elevates validity to an explicit research object and develops a two-level validity assessment framework for RL-ABS. At the single-agent level, the key question is whether the adopted bid representation and learning algorithm can approach the theoretical optimal profit under a prescribed environment; otherwise, multi-agent mechanism analysis would lack the most basic foundation of learning correctness. At the multi-agent level, the key question is whether the stabilized policy profile is close to a state in which unilateral deviations cannot yield significant profit improvement, thereby indicating its distance from the equilibrium region. The former answers whether the model has learned correctly, whereas the latter answers whether the resulting policy profile is sufficiently reliable to support mechanism comparison. Only when both levels are satisfactorily addressed can RL-ABS be elevated from a runnable simulation tool to a reliable research instrument for electricity market mechanism analysis.

\subsection{Research logic}

Against the above background, this paper is organized around two fundamental issues in continuous multi-segment bid learning for electricity markets. The first is how to construct a bid representation that is consistent with policy-gradient optimization while satisfying market-rule constraints. The second is how to assess whether RL-ABS outputs already possess the validity required for mechanism analysis. The former corresponds to the representation problem in constrained continuous action learning, whereas the latter corresponds to the problem of result interpretation in electricity market simulation. Together, these two issues constitute the central logic of this paper from the two perspectives of whether the learning object is correctly specified and whether the research conclusions are trustworthy.

Along the first line of inquiry, this paper first analyzes the necessary conditions that a post-processing mapping should satisfy under the policy gradient framework, and then reveals the theoretical limitations of common constrained post-processing operations such as sorting, clipping, and projection. On this basis, DPMP is proposed to enable the continuous generation of feasible multi-segment bids. Along the second line of inquiry, this paper further establishes a two-level validity assessment framework for RL-ABS, evaluating the validity of simulation results from two perspectives: proximity to the theoretical optimum in the single-agent setting and equilibrium stability in the multi-agent setting. Accordingly, the paper forms an integrated research chain of problem definition, theoretical diagnosis, parametric method design, validity assessment, and experimental verification.

\section{Necessary Conditions for Post-Processing Mappings under Policy Gradient Methods and Theoretical Limitations of Common Constraint-Handling Approache}

This section formally characterizes, within a unified framework, the relationship among the raw outputs of the policy network, the post-processing mapping, and the feasible bids ultimately submitted to the market. Since both the rewards and the state transitions of the environment are determined by the feasible bids that are actually executed in the market, rather than by the raw outputs of the policy network, the policy gradient update should remain consistent with the distribution induced by the executed actions. On this basis, this paper first derives the necessary conditions that a post-processing mapping must satisfy under the policy gradient framework, and then analyzes why common operations such as sorting, clipping, and projection may violate these conditions.

\subsection{Necessary Conditions for Post-Processing Mappings under Policy Gradient Methods}

\subsubsection{Objective Function and Basic Gradient Form of Stochastic Policy Gradient Methods}

In single-agent reinforcement learning, let the stochastic policy be denoted by $\mu_\theta(\cdot \mid s)$, and let the objective be to maximize the expected discounted return:

\begin{equation}
J(\theta) \coloneqq \mathbb{E}_{\tau \sim \mu_\theta}
\left[
\sum_{t \ge 0} \gamma^t r(s_t,a_t)
\right]
\label{eq:pg_objective}
\end{equation}

Under standard regularity conditions, the policy gradient theorem gives \cite{sutton1998reinforcement}\cite{zhao2025mathematical}:

\begin{equation}
\nabla_\theta J(\theta)
=
\mathbb{E}_{\tau \sim \mu_\theta}
\left[
\sum_{t \ge 0}
\nabla_\theta \log \mu_\theta(a_t \mid s_t)\, G_t
\right]
\label{eq:policy_gradient}
\end{equation}

where $G_t$ denotes the return or an estimate of the advantage function. A key premise implicit in this expression is that the action variable appearing in the gradient must coincide with the action that is actually executed by the environment and generates the reward.

\subsubsection{Executed-Action Distribution Induced by the Post-Processing Mapping under Stochastic Policy Gradient Methods}

In electricity market bidding problems, the policy network often does not directly output a feasible bid $a \in \mathcal{A}$ that satisfies all constraints. Instead, it first outputs an unconstrained (or weakly constrained) vector $x \in \mathbb{R}^d$, which is then transformed into a feasible action through a deterministic post-processing mapping $h:\mathbb{R}^d \to \mathcal{A}$, yielding

\begin{equation}
a = h(x)
\label{eq:post_mapping}
\end{equation}

Since the environment and the reward function depend only on the final submitted action $a$, rather than on $x$, the true optimization objective of policy learning must be defined on the basis of the trajectory distribution induced by the executed action $a$.

Let $\mu_\theta(\cdot \mid s)$ denote the conditional distribution of the raw output $x$ under state $s$. Then the executed-action distribution $\pi_\theta(\cdot \mid s)$ is defined as its pushforward distribution under the mapping $h$:

\begin{equation}
\pi_\theta(\cdot \mid s) = h_{\#}\mu_\theta(\cdot \mid s)
\label{eq:pushforward_policy}
\end{equation}

We now embed the above one-step relation into a sequential decision process
(an MDP, or equivalently a Markov game viewed from the perspective of a single agent). At each time step $t$, the process proceeds as follows:

\begin{enumerate}[label=\roman*.]
    \item observe the state $s_t$;
    \item sample a raw action $x_t \sim \mu_\theta(\cdot \mid s_t)$;
    \item generate a feasible action $a_t = h(x_t)$ and submit it to the market;
    \item the environment generates the next state $s_{t+1}$ according to $P(\cdot \mid s_t,a_t)$, and returns the reward $r(s_t,a_t)$.
\end{enumerate}

Therefore, given an initial distribution $\rho_0$, the trajectory distribution induced by this decision system can be written as

\begin{equation}
p_\theta(\tau)
=
\rho_0(s_0)
\prod_{t\ge 0}
\Bigl[
\mu_\theta(x_t \mid s_t)\cdot
\mathbf{1}\{a_t=h(x_t)\}\cdot
P(s_{t+1}\mid s_t,a_t)
\Bigr]
\label{eq:traj_raw}
\end{equation}

Furthermore,

\begin{equation}
p_\theta(\tau_a)
=
\rho_0(s_0)
\prod_{t\ge 0}
\Bigl[
\pi_\theta(a_t\mid s_t)\cdot
P(s_{t+1}\mid s_t,a_t)
\Bigr],
\qquad
\pi_\theta(\cdot\mid s)=h_{\#}\mu_\theta(\cdot\mid s)
\label{eq:traj_exec}
\end{equation}

where $\tau_a=(s_0,a_0,s_1,a_1,\ldots)$ denotes the trajectory actually experienced by the environment. Accordingly, the true optimization objective should be written as

\begin{equation}
J(\theta)
=
\mathbb{E}_{\tau_a\sim p_\theta}
\left[
\sum_{t\ge 0}\gamma^t r(s_t,a_t)
\right]
=
\mathbb{E}
\left[
\sum_{t\ge 0}\gamma^t r\bigl(s_t,h(x_t)\bigr)
\right]
\label{eq:true_objective}
\end{equation}

Therefore, any policy gradient update for optimizing $J(\theta)$ must, in principle, be constructed around $\pi_\theta$, rather than tacitly retaining the form of $\mu_\theta$.

\subsubsection{Necessary Conditions for Post-Processing Mappings under Stochastic Policy Gradient Methods}

Since the update terms in stochastic policy gradient methods (e.g., REINFORCE, QAC, A2C, TRPO, and PPO) all require the computation of

\begin{equation}
\nabla_\theta \log \pi_\theta(a \mid s)
\label{eq:score_exec}
\end{equation}

where $\pi_\theta(\cdot \mid s)$ is the true executed-action distribution induced by $a=h(x)$. In order for $\log \pi_\theta(a \mid s)$ and its gradient to be well-defined, estimable, and amenable to stable backpropagation, several necessary conditions must be satisfied. To better investigate these necessary conditions, we first introduce the concepts of a non-redundant action space and a non-redundant mapping.

Definition 3.1 (Non-redundant action space and non-redundant mapping $g$). Let $z=\phi(x)\in\mathbb{R}^{d_z}$ denote the coordinate of the raw stochastic policy output $x$ in a non-redundant action space with respect to the post-processing mapping $h$. If there exist mappings

\begin{equation}
\phi:\mathbb{R}^{d_x}\to\mathbb{R}^{d_z},
\qquad
g:\mathbb{R}^{d_z}\to\mathcal{A}
\label{eq:nonredundant_maps}
\end{equation}

such that, for the raw stochastic policy output $x$,

\begin{equation}
h(x)=g\bigl(\phi(x)\bigr)
\label{eq:h_factorization}
\end{equation}

Then $g$ is called a non-redundant mapping. The motivation for introducing a non-redundant action space is that the output of the raw stochastic policy $x$ is often redundant. For example, when a generating unit outputs the power quantities of $K$ segments, any one segment can be recovered by subtracting the other $K-1$ segment quantities from the total generation quantity. Hence, one dimension is redundant. In the subsequent discussion of NC2 and NC3, the main objects of interest are the non-redundant action space $z$ and the non-redundant mapping $g$.

The three necessary conditions for the post-processing mapping are given below, corresponding respectively to requirements at the distributional level, the global mapping level, and the local differential level: NC1 excludes singular probability mass, NC2 excludes branch ambiguity, and NC3 excludes local gradient collapse. Detailed proofs are provided in Appendix A.

i. Necessary Condition 1 (NC1): The post-processing action mapping $h$ should satisfy
\begin{equation}
\forall a_0,\ \mathbb{P}\bigl(h(X)=a_0 \mid s\bigr)=0
\label{eq:nc1}
\end{equation}

This condition means that the true executed-action distribution must not assign positive probability mass to isolated points or low-dimensional constraint manifolds.

ii. Necessary Condition 2 (NC2): The non-redundant mapping $g$ of the post-processing action must be injective, i.e.,

\begin{equation}
\mathbb{P}\bigl(\exists z' \neq z:\ g(z')=g(z)\mid s\bigr)=0
\label{eq:nc2}
\end{equation}

where $s$ denotes the state space. This condition means that, on the non-redundant action space, the post-processing mapping cannot be globally many-to-one.

iii. Necessary Condition 3 (NC3): The non-redundant mapping $g$ of the post-processing action must be locally invertible on the non-redundant action space $z$, i.e.,

\begin{equation}
\mathbb{P}\bigl(\det J_g(z)\neq 0 \mid s\bigr)=1
\label{eq:nc3}
\end{equation}

where $J_g$ denotes the Jacobian matrix of the mapping $g$.

\subsubsection{Corresponding Conditions under Deterministic Policy Gradient Methods}

Similarly, in electricity market bidding, a deterministic policy typically outputs a raw action $x\in\mathbb{R}^d$ through the policy network, which is then transformed into a feasible action through a deterministic post-processing mapping $h:\mathbb{R}^d\to\mathcal{A}$:

\begin{equation}
a=h(x), \qquad x=f_\theta(s)
\label{eq:det_postprocess}
\end{equation}

Therefore, the true executed policy is not $f_\theta$, but rather the composite mapping

\begin{equation}
\tilde{\mu}_\theta(s)=h\bigl(f_\theta(s)\bigr)
\label{eq:true_det_policy}
\end{equation}

Here, $\tilde{\mu}_\theta$ denotes the post-processed deterministic executed policy. Under standard regularity conditions, the deterministic policy gradient theorem gives\cite{sutton1998reinforcement}\cite{zhao2025mathematical},

\begin{equation}
\nabla_\theta J(\theta)
=
\mathbb{E}_{s\sim \rho^{\tilde{\mu}_\theta}}
\left[
\nabla_\theta \tilde{\mu}_\theta(s)\,
\nabla_a Q^{\tilde{\mu}_\theta}(s,a)\big|_{a=\tilde{\mu}_\theta(s)}
\right]
\label{eq:dpg_true}
\end{equation}

Since $\tilde{\mu}_\theta=h\circ f_\theta$, the key chain-rule decomposition is

\begin{equation}
\nabla_\theta \tilde{\mu}_\theta(s)
=
\left. J_h(x)\right|_{x=f_\theta(s)}\,\nabla_\theta f_\theta(s)
\label{eq:chain_rule_h}
\end{equation}

where $J_h$ denotes the Jacobian matrix of $h$. In algorithmic implementations, deterministic policies typically introduce exploration through behavior-policy noise or target-policy smoothing. More generally, this can be written as

\begin{equation}
x=f_\theta(s)+\sigma\varepsilon,\qquad
\varepsilon\sim \varphi(\cdot),\qquad
a=h(x),
\label{eq:det_noise_postprocess}
\end{equation}

Accordingly, for a given state $s$, let the conditional distribution of $x$ be denoted by $\nu_{\theta,\sigma}(\cdot\mid s)$. Then the conditional distribution of the executed action is its pushforward distribution under $h$:

\begin{equation}
\pi_{\theta,\sigma}(\cdot\mid s)=h_{\#}\nu_{\theta,\sigma}(\cdot\mid s)
\label{eq:det_pushforward}
\end{equation}

As $\sigma\to 0$, we have $\nu_{\theta,\sigma}(\cdot\mid s)\Rightarrow \delta_{f_\theta(s)}$, and hence $\pi_{\theta,\sigma}(\cdot\mid s)\Rightarrow \delta_{\tilde{\mu}_\theta(s)}$. This shows that a deterministic policy is the limiting case of a family of noisy randomized executed policies $\{\pi_{\theta,\sigma}\}_{\sigma>0}$. If one requires the algorithm to preserve consistency between the true execution objective and the gradient update for arbitrarily small $\sigma$, then the constraints imposed on $h$ must be structurally identical to those imposed on the pushforward distribution in the stochastic policy gradient setting, namely, NC1--NC3.

\subsection{Theoretical Limitations of Three Common Post-Processing Approaches under Policy Gradient Methods}

The following subsections show, respectively, that sorting primarily destroys injectivity (NC2), clipping mainly introduces singular probability mass on the boundary or lower-dimensional constraint manifolds (NC1), and projection simultaneously induces many-to-one mappings and local degeneracy (NC2–NC3).

\subsubsection{Deficiencies of Sorting as a Post-Processing Operation under Policy Gradient Methods}

\begin{figure}[H]
    \centering
    \includegraphics[width=\linewidth]{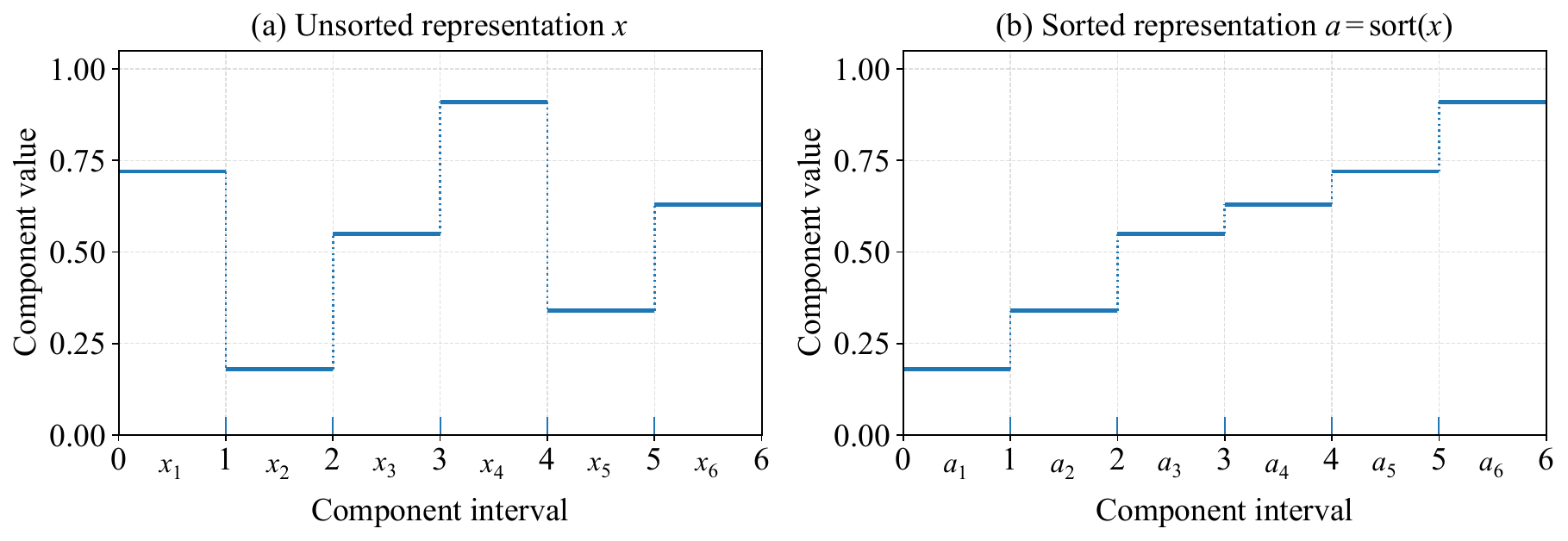}
    \caption{Conceptual illustration of sorting.}
    \label{fig:sorting}
\end{figure}

The sorting post-processing operation sorts the price output $x\in\mathbb{R}^K$ of the raw policy network in ascending order, as illustrated in Figure~\ref{fig:sorting}:

\begin{equation}
a=h(x)=\operatorname{sort}(x)
\label{eq:sort_mapping}
\end{equation}

If $x$ is a continuous random vector, then with probability one, $x_i\neq x_j$ for $i\neq j$. For any such $x$, let $a=\operatorname{sort}(x)$. Then, for any permutation $\sigma\in S_K$, we have

\begin{equation}
\operatorname{sort}(P_\sigma x)=a
\label{eq:sort_perm_invariance}
\end{equation}

Moreover, whenever all components $x_i$ are pairwise distinct, the permuted vectors $P_\sigma x$ are mutually different. Therefore, for almost every reachable $a$, its preimage contains at least $K!$ distinct points:

\begin{equation}
\left|h^{-1}(a)\right|\ge K!
\label{eq:sort_preimage}
\end{equation}

Therefore, the sorting post-processing mapping violates NC2, thereby causing an objective mismatch in stochastic policy gradient optimization. More specifically, the sorting operation compresses raw outputs with different segment identities into the same sorted result, so that the branch information linking the executed action to the raw output becomes irrecoverable. For multi-segment bids, this further implies that positional information with original economic meaning—such as the price of the j-th segment—is erased.

\subsubsection{Deficiencies of clipping-based post-processing under policy gradient methods}

To satisfy interval constraints or monotonicity constraints, two common types of clipping-based post-processing mappings can be employed.

\noindent
(1) Constant-bound clipping applied to $x$:

\begin{figure}[H]
    \centering
    \includegraphics[width=\linewidth]{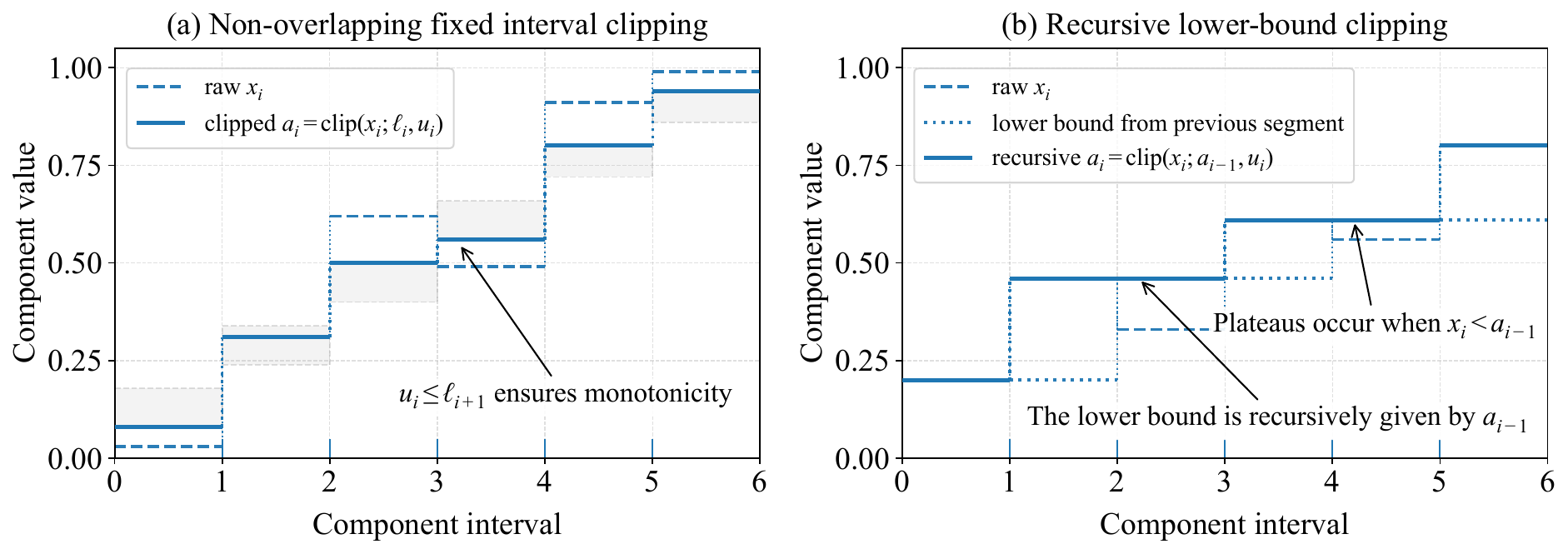}
    \caption{Conceptual illustrations of two clipping-based post-processing schemes.}
    \label{fig:clip}
\end{figure}

As illustrated in Fig.~\ref{fig:clip}(a), if mutually compatible fixed lower and upper bounds are pre-specified for each segment, then the executed bids can be obtained by segment-wise constant-bound clipping so as to satisfy the bound constraints and, under particular settings, the monotonicity requirement:

\begin{equation}
a_i=\operatorname{clip}(x_i;\ell_i,u_i)=
\begin{cases}
\ell_i, & x_i<\ell_i,\\
x_i, & \ell_i\le x_i\le u_i,\\
u_i, & x_i>u_i.
\end{cases}
\label{eq:constant_clip}
\end{equation}

If $\mathbb{P}(x_i>u_i\mid s)>0$ (or $\mathbb{P}(x_i<\ell_i\mid s)>0$), then

\begin{equation}
\mathbb{P}(a_i=u_i\mid s)=\mathbb{P}(x_i>u_i\mid s)>0
\qquad\text{or}\qquad
\mathbb{P}(a_i=\ell_i\mid s)>0
\label{eq:clip_atom}
\end{equation}

Hence, the executed-action distribution places positive probability mass on the boundary points and therefore cannot admit a purely continuous density, violating NC1. The issue is not merely that the values of xare truncated; rather, an entire region of the input space is collapsed onto the same boundary point.

\noindent
(2) Recursive-bound clipping applied to $x$:

In many implementations, clipping is not performed against fixed constant bounds. Instead, outputs are generated sequentially, with the previous segment taken as the lower or upper bound so as to enforce monotonicity. For example, as illustrated in Fig.~\ref{fig:clip}(b), under a monotone nondecreasing constraint, the output can be written as

\begin{equation}
a_1=x_1,\qquad
a_i=\max(a_{i-1},x_i),\quad (i\ge 2)
\label{eq:recursive_clip_max}
\end{equation}

More generally,

\begin{equation}
a_i=\operatorname{clip}(x_i;\,a_{i-1},u_i),\quad (i\ge 2)
\label{eq:recursive_clip_general}
\end{equation}

Taking Eq.~\eqref{eq:recursive_clip_max} as an example, by definition,

\begin{equation}
a_i=a_{i-1}
\iff
x_i\le a_{i-1}
\label{eq:recursive_atom_condition}
\end{equation}

Therefore, for any given $a_{i-1}$,
\begin{equation}
\mathbb{P}(a_i=a_{i-1}\mid a_{i-1},s)
=
\mathbb{P}(x_i\le a_{i-1}\mid a_{i-1},s)
\label{eq:recursive_equal_cond}
\end{equation}

Under continuously distributed outputs, the right-hand side is typically strictly positive. Taking expectation with respect to $a_{i-1}$ yields

\begin{equation}
\mathbb{P}(a_i=a_{i-1}\mid s)>0
\label{eq:recursive_equal_positive}
\end{equation}

Therefore, the joint executed-action distribution $\pi_\theta(a\mid s)$ assigns positive probability mass to the set $\{a_i=a_{i-1}\}$, and thus is not a purely continuous density on $\mathbb{R}^K$, violating NC1.

In summary, whether the clipping bounds are constant or recursively determined by the previous segment to enforce monotonicity, the common feature is that a continuous region of the input space is collapsed onto boundary points or equality-constraint surfaces, so that the executed-action distribution no longer satisfies the regular probability density function required by NC1.

\subsubsection{Deficiencies of projection-based post-processing under policy gradient methods}

\begin{figure}[H]
    \centering
    \includegraphics[width=0.55\linewidth]{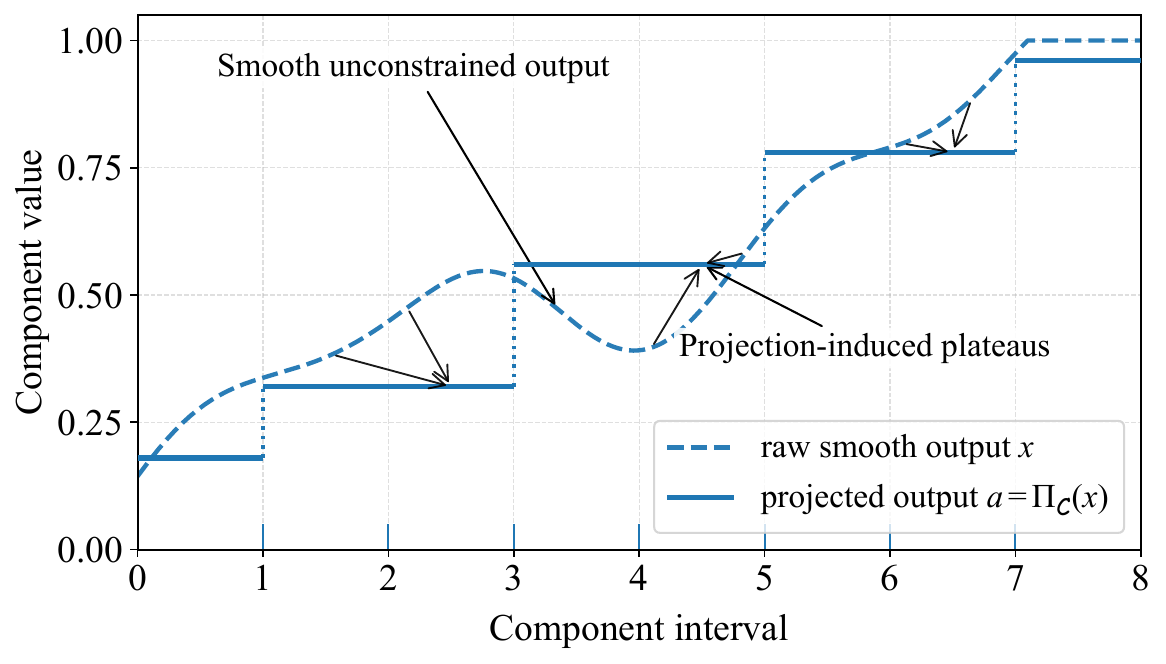}
    \caption{Conceptual illustration of projection-induced staircase behavior.}
    \label{fig:projection}
\end{figure}

When the policy network outputs a continuous but irregular bid curve, projection-based post-processing is often considered to enforce the bid-curve constraints. As illustrated in Fig.~\ref{fig:projection}, a common approach is to use Euclidean projection

\begin{equation}
a=\Pi_C(x):=\arg\min_{y\in C}\|y-x\|_2^2
\label{eq:euclidean_projection}
\end{equation}

where $C\subset\mathbb{R}^d$ denotes the feasible set that is closed, convex, monotone, and subject to upper and lower bounds. Take any $a\in\partial C$ at which active constraints are present. For Euclidean projection onto a closed convex set, if $a$ lies on the boundary and active constraints exist, then within a sufficiently small neighborhood along an outward normal direction, all points are projected onto the same boundary point $a$. That is, there exist a nonzero direction $n$ and some $\tau_0>0$ such that, for all $\tau\in(0,\tau_0)$,

\begin{equation}
\Pi_C(a+\tau n)=a
\label{eq:projection_flat}
\end{equation}

Hence, infinitely many distinct $x$ are mapped to the same $a$; that is, the projection is many-to-one at the boundary, violating NC2. In addition, Eq.~\eqref{eq:projection_flat} implies that the first-order variation of the output along direction $n$ is zero. Therefore, in the differentiable sense (or almost everywhere in the piecewise differentiable sense), there exists a nonzero vector $v$ such that

\begin{equation}
J_{\Pi_C}(x)\,v=0
\label{eq:projection_jacobian_singular}
\end{equation}

which indicates that the Jacobian cannot be full rank, thereby violating NC3.

\section{Dual-Positive Monotone Parameterization for Stepwise Bids}

\begin{figure}[H]
    \centering
    \includegraphics[width=0.75\linewidth]{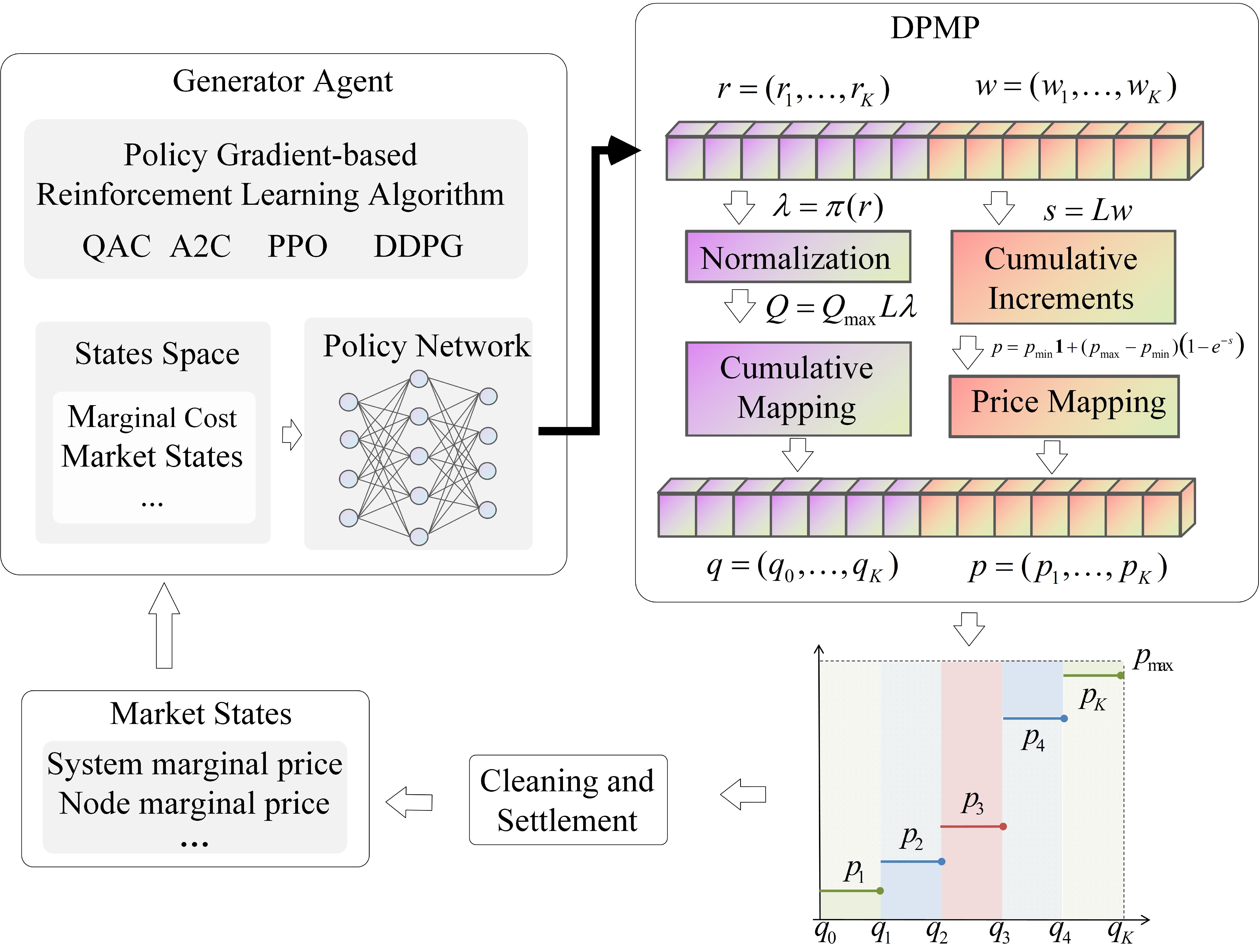}
    \caption{DPMP-based framework for stepwise bid generation of a generator agent.}
    \label{fig:Algorithm Explanation Diagram}
\end{figure}

To avoid the disruption of policy-gradient consistency caused by post-processing operations such as sorting, clipping, and projection, this paper proposes the Dual-Positive Monotone Parameterization (DPMP). As shown in Figure~\ref{fig:Algorithm Explanation Diagram}, the core idea is to let the policy network directly output two positive vectors, which respectively represent the segment-wise generation-width and the price-increment, and then to construct, via a continuously differentiable mapping, a multi-segment stepwise bid curve that satisfies the required constraints.

\subsection{Feasible Set of Stepwise Bid Curves}

A $K$-segment stepwise bid curve is determined by the generation output breakpoints and the segment prices:

\begin{equation}
0=Q_0<Q_1<\cdots<Q_{K-1}<Q_K=Q_{\max},\qquad
p_1\le p_2\le\cdots\le p_K
\label{eq:step_bid_basic}
\end{equation}

where the constant bid price over the generation interval $[Q_{i-1},Q_i]$ is denoted by $p_i$. Let the set of all curves satisfying the above constraints be denoted by

\begin{equation}
\mathcal{B}_K
=
\left\{
(Q,p)\ \middle|\
0=Q_0<Q_1<\cdots<Q_K=Q_{\max},\;
p_{\min}\le p_1\le\cdots\le p_K\le p_{\max}
\right\}
\label{eq:BK_closed}
\end{equation}

It should be noted that the closed feasible set $\mathcal{B}_K$ allows plateaus between adjacent price segments, i.e., $p_i=p_{i+1}$. To achieve better differentiability, injectivity, and local invertibility, this paper does not directly parameterize the entire feasible set; instead, it parameterizes its interior subset:

\begin{equation}
\mathcal{B}_K^\circ
=
\left\{
(Q,p)\ \middle|\
0=Q_0<Q_1<\cdots<Q_K=Q_{\max},\;
p_{\min}<p_1<\cdots<p_K<p_{\max}
\right\}
\label{eq:BK_open}
\end{equation}

This interior feasible region corresponds to multi-segment bid curves that are strictly increasing and strictly bounded. From an engineering implementation perspective, $\mathcal{B}_K^\circ$ already covers the feasible action space required in continuous bidding learning. Meanwhile, every boundary point of $\mathcal{B}_K$ can be approximated arbitrarily closely by a sequence in $\mathcal{B}_K^\circ$. Therefore, this restriction does not weaken the model's expressive capability with respect to bid sets observed in actual electricity markets.

\subsection{Action space}

Define the action space output by the policy network as:

\begin{equation}
a=(r_1,\ldots,r_K,\;w_1,\ldots,w_K)\in \mathbb{R}_{>0}^{2K}
\label{eq:dual_positive_action}
\end{equation}

where $r_i$ denotes the generation-output width parameter of segment $i$, and $w_i$ denotes the price-increment parameter of segment $i$. A feasible stepwise bid curve can then be obtained from this action vector through a one-step mapping. In addition, both $r_i$ and $w_i$ are required to satisfy the dual-positive condition. Here, “dual-positive” means that the action vector is divided into two parts: the generation-output component $r=(r_1,\ldots,r_K)$, whose entries are all strictly positive, and the price component $w=(w_1,\ldots,w_K)$, whose entries are likewise all strictly positive. Hence, the entire action vector lies in a positive orthant. This dual-positive design has two advantages. First, both the positive generation-output width parameters and the positive price-increment parameters can be generated stably by common activation functions such as Softplus, Exp, or Sigmoid, which facilitates
practical implementation. Second, the positivity constraints are directly absorbed into the parameterization structure itself, thereby ruling out, at the source, illegal actions such as negative output widths, negative increments, and non-monotone prices.

\subsection{Generation-output mapping}

The generation-output mapping is carried out in two steps. First, the width parameters are normalized as $\lambda=\pi(r)$:

\begin{equation}
\lambda_i=\frac{r_i}{\sum_{j=1}^{K} r_j},
\qquad
\sum_{i=1}^{K}\lambda_i=1,\quad
\lambda_i>0
\label{eq:lambda_normalization}
\end{equation}

Next, define the lower-triangular accumulation operator $L\in\mathbb{R}^{K\times K}$:

\begin{equation}
L=
\begin{bmatrix}
1 & 0 & \cdots & 0\\
1 & 1 & \ddots & \vdots\\
\vdots & \ddots & \ddots & 0\\
1 & \cdots & 1 & 1
\end{bmatrix}
\label{eq:accumulation_operator}
\end{equation}

A further cumulative mapping then yields the generation output breakpoints:

\begin{equation}
Q=Q_{\max}L\lambda
\label{eq:Q_from_lambda}
\end{equation}

that is,

\begin{equation}
Q_i=\left(\sum_{j=1}^{i}\lambda_j\right)Q_{\max},
\qquad
Q_0=0
\label{eq:Qi_cumsum}
\end{equation}

Since $\lambda_i>0$ and $\sum_{i=1}^{K}\lambda_i=1$, it follows immediately that

\begin{equation}
0=Q_0<Q_1<\cdots<Q_K=Q_{\max}
\label{eq:Q_strict_increasing}
\end{equation}

Therefore, the above mapping naturally guarantees that the generation output breakpoints are strictly increasing, with the terminal point fixed at the unit's maximum generation output $Q_{\max}$. Moreover, this mapping also has a clear inverse structure. Given any breakpoint vector $Q$ satisfying the constraints, $\lambda$ can be uniquely recovered by first-order differencing:

\begin{equation}
\lambda_1=\frac{Q_1}{Q_{\max}},
\qquad
\lambda_i=\frac{Q_i-Q_{i-1}}{Q_{\max}},\quad i=2,\ldots,K
\label{eq:lambda_inverse}
\end{equation}

This provides the basis for the subsequent proof of injectivity and local invertibility of the mapping.

\subsection{Price mapping}

To simultaneously satisfy the requirements of strictly increasing prices and upper–lower bound constraints, this paper does not adopt a simple linear cumulative form. Instead, we first define the cumulative increments:

\begin{equation}
s_i=\sum_{j=1}^{i}w_j,\qquad i=1,\ldots,K
\label{eq:s_cumsum}
\end{equation}

that is,
\begin{equation}
s=Lw
\label{eq:s_vector_form}
\end{equation}

We then map $s$ into the price interval $(p_{\min},p_{\max})$ by
\begin{equation}
p
=
p_{\min}\mathbf{1}
+
(p_{\max}-p_{\min})\bigl(1-e^{-s}\bigr)
\label{eq:p_from_s}
\end{equation}

where the exponential function is applied elementwise. This mapping has the following three properties.

First, since $w_i>0$, we have $s_i>0$, and hence

\begin{equation}
0<1-e^{-s_i}<1
\label{eq:exp_range}
\end{equation}

Therefore,

\begin{equation}
p_{\min}<p_i<p_{\max},\qquad i=1,\ldots,K
\label{eq:p_bounds}
\end{equation}

This shows that the prices are naturally confined within the prescribed bounds.

Second, since $w_i>0$, it follows that

\begin{equation}
s_1<s_2<\cdots<s_K
\label{eq:s_strict_increasing}
\end{equation}

Because the function $x\mapsto 1-e^{-x}$ is strictly increasing on $(0,+\infty)$, we obtain

\begin{equation}
p_1<p_2<\cdots<p_K
\label{eq:p_strict_increasing}
\end{equation}

That is, the price curve is strictly monotonically increasing.

Third, this mapping is invertible. Given any price vector
$p\in(p_{\min},p_{\max})^K$, one may first recover
\begin{equation}
s_i
=
-\ln\!\left(
1-\frac{p_i-p_{\min}}{p_{\max}-p_{\min}}
\right),
\qquad i=1,\ldots,K
\label{eq:s_from_p}
\end{equation}

and then obtain, by differencing,
\begin{equation}
w_1=s_1,\qquad
w_i=s_i-s_{i-1},\qquad i=2,\ldots,K
\label{eq:w_from_s}
\end{equation}

Hence, there exists a one-to-one correspondence between the price-increment parameter $w$ and the feasible price vector $p$. Compared with post-processing approaches such as sorting, clipping, and projection, the above price mapping neither compresses different inputs into the same output nor creates kinks or extended flat regions on the constraint boundaries. It is therefore better aligned with the requirement of continuous differentiability in action mappings for policy gradient methods.

\subsection{DPMP mapping}

Let $\mathcal{A}$ denote the action space, i.e.,

\begin{equation}
\mathcal{A}:=\mathbb{R}_{>0}^{2K}
=
\underbrace{\mathbb{R}_{>0}^{K}}_{r}
\times
\underbrace{\mathbb{R}_{>0}^{K}}_{w}
\label{eq:action_space_dpmp}
\end{equation}

The DPMP mapping $\Phi:\mathcal{A}\to\mathcal{B}_K^\circ$ is defined as

\begin{equation}
Q=Q_{\max}L\lambda(r),
\qquad
p=p_{\min}\mathbf{1}+(p_{\max}-p_{\min})\bigl(1-e^{-Lw}\bigr)
\label{eq:dpmp_mapping}
\end{equation}

As established in the preceding two subsections, DPMP stably maps any action in the positive orthant to an interior-feasible monotone multi-segment bid curve. One further point deserves emphasis: the normalization mapping in the generation-output component depends only on the direction of $r$, but not on its positive scalar magnitude. That is, for any $c>0$,

\begin{equation}
\pi(cr)=\pi(r)
\label{eq:scale_invariance_pi}
\end{equation}

Therefore, the raw variable $r$ contains a one-dimensional scale redundancy. After removing this redundancy, DPMP yields a smooth injective mapping from the non-redundant action space to the interior feasible bid domain, while retaining local invertibility. Appendix B further proves, in non-redundant coordinates, that DPMP satisfies the necessary conditions for post-processing operations proposed in this paper, namely NC1–NC3. In other words, unlike the SORT/CLIP/PROJECT baselines, it does not introduce singular mass on the boundary, branch ambiguity, or local gradient collapse.

\section{Validity Assessment Framework for Electricity Market RL-ABS}

\begin{figure}[H]
    \centering
    \includegraphics[width=0.75\linewidth]{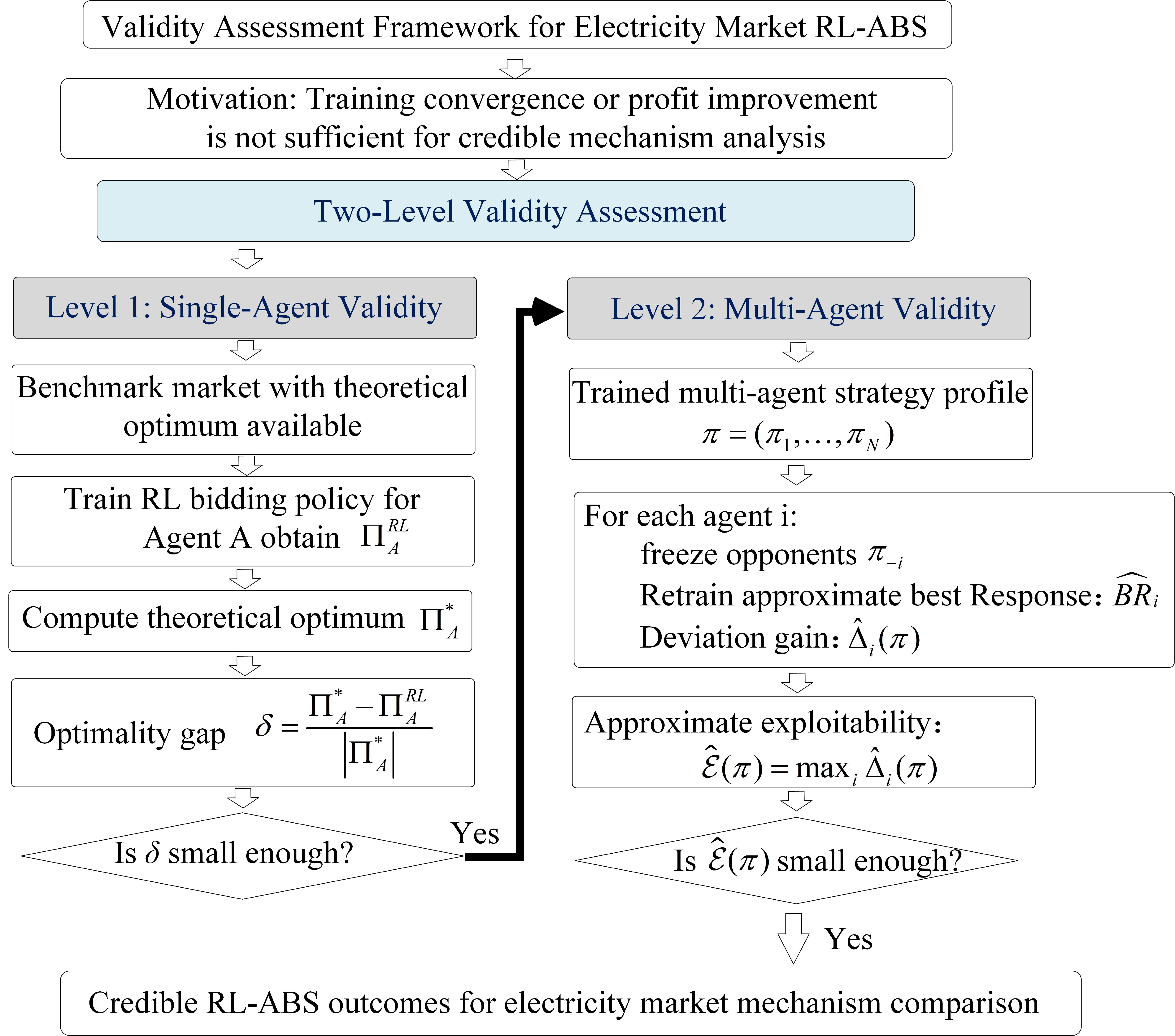}
    \caption{Validity Assessment Framework for Electricity Market RL-ABS.}
    \label{fig:Validity Assessment Framework for Electricity Market RL-ABS}
\end{figure}

As discussed earlier, training convergence or profit improvement alone is insufficient to demonstrate that RL-ABS results can be directly used for electricity market mechanism analysis and evaluation. This chapter therefore develops a two-level validity assessment framework for electricity market RL-ABS to verify the credibility of learned policies before conducting large-scale mechanism evaluation, as shown in Figure~\ref{fig:Validity Assessment Framework for Electricity Market RL-ABS}. Specifically, in a single-agent setting with a theoretical optimal profit benchmark, the optimality gap is used to assess how closely the RL policy approximates the optimal bidding behavior. Subsequently, in a multi-agent general-sum game, exploitability based on approximate best responses is employed to measure the robustness of a policy profile against unilateral deviations, thereby evaluating its proximity to the ε-Nash equilibrium. This framework provides quantifiable and reproducible validity evidence for applying RL-ABS to electricity market mechanism comparison.

\subsection{Single-Agent Algorithm Validity Assessment}

This section addresses a fundamental yet often overlooked question: when reinforcement learning is applied to electricity market agent-based simulation (ABS), does it truly learn the optimal bidding strategy? To avoid misleading conclusions drawn solely from convergence curves or higher profits, the validity assessment of RL strategies in electricity market ABS requires, at this stage, a benchmark setting in which the theoretical optimal bidding strategy is available. On this basis, the validity of single-agent learning outcomes is evaluated by comparing the profit gap between the RL-learned strategy and the theoretical optimal strategy.

\noindent
\textbf{(1) Benchmark market setting and bid structure}

Consider a single-node day-ahead market with $96$ time intervals under a uniform-price clearing mechanism, involving two generators: the learning generator $A$ and the rival generator $B$. The system demand is denoted by $d(t)$. At each time interval, after both participants submit their supply bids, the Independent System Operator (ISO) clears the market by accepting bids in ascending order until demand is satisfied, and determines the uniform clearing price $\lambda$ according to the marginal accepted bid. Generator $B$ follows cost-based bidding behavior, i.e., its bid is constructed as a stepwise curve from its own marginal cost function. The learning generator $A$ submits a ten-segment stepwise bid, a format widely adopted in real-world electricity markets.

\noindent
\textbf{(2) Total cost and marginal cost model \cite{pan2025decision}}

To avoid the systematic bias introduced by linearized cost assumptions in the low-output operating region, the total cost of generator $g$ is modeled using a polynomial function:

\begin{equation}
C_g^T(p_g)
=
\beta_{g,0}
+
\sum_{j=1}^{d}\beta_{g,j}p_g^{\,j},
\qquad
g\in\{A,B\}
\label{eq:total_cost_poly}
\end{equation}

Its marginal cost is obtained by differentiation:

\begin{equation}
C_g^M(p_g)
=
\frac{d\,C_g^T(p_g)}{d p_g}
=
\sum_{j=1}^{d} j\,\beta_{g,j}\,p_g^{\,j-1}
\label{eq:marginal_cost_poly}
\end{equation}

where $d$ denotes the polynomial order. Prior studies have shown that polynomial functions, particularly with $d=2$ or $3$, can better capture the nonlinear characteristics of marginal cost when a generating unit operates under low-load conditions.

\noindent
\textbf{(3) Theoretical optimal profit}

Based on the supply curve submitted by generator $B$, for any given price $\lambda$, its supplied energy can be written as

\begin{equation}
S_B(\lambda)
=
\operatorname{clip}\!\left((C_B^M)^{-1}(\lambda),\,0,\,\bar{P}_B\right)
\label{eq:SB_lambda}
\end{equation}

where $\bar{P}_B$ denotes the upper bound of generation output for generator $B$, and $\operatorname{clip}(\cdot)$ denotes clipping onto the feasible interval $[0,\bar{P}_B]$. At a given time interval, let the demand be $D$. When the uniform clearing price is $\lambda$, the residual demand faced by generator $A$ after deducting B's supplied quantity is

\begin{equation}
R(\lambda)
=
\operatorname{clip}\!\left(D-S_B(\lambda),\,0,\,\bar{P}_A\right)
\label{eq:R_lambda}
\end{equation}

where $\bar{P}_A$ is the upper bound of generation output for generator $A$. If demand uncertainty is taken into account, $D$ can be treated as a random variable and the following expressions can be evaluated in the expectation sense. Accordingly, the corresponding profit of generator $A$ is

\begin{equation}
\Pi_A(\lambda)=\lambda R(\lambda)-C_A^T\!\bigl(R(\lambda)\bigr)
\label{eq:profit_A_lambda}
\end{equation}

Therefore, the theoretical optimal profit is given by the maximum value over the feasible price set:

\begin{equation}
\Pi_A^*
=
\max_{\lambda\in\Omega}
\left[
\lambda R(\lambda)-C_A^T\!\bigl(R(\lambda)\bigr)
\right]
\label{eq:profit_A_star}
\end{equation}

\noindent
\textbf{(4) Validity metric and statistical criterion}

Define the optimality gap as

\begin{equation}
\delta
=
\frac{\Pi_A^*-\Pi_A^{RL}}{\left|\Pi_A^*\right|}
\label{eq:optimality_gap}
\end{equation}

where $\Pi_A^{RL}$ denotes the profit achieved by the reinforcement learning policy. When the upper bound of the confidence interval of $\delta$ is below a prescribed threshold $\tau$, the single-agent algorithm is considered to have passed the validity assessment under this verifiable benchmark.

\subsection{Multi-agent validity assessment in RL-ABS}
In multi-agent electricity markets, merely observing convergence of the training curves or improvements in profit does not demonstrate that the learned policies are stable in the game-theoretic sense, and therefore cannot serve as a credible basis for mechanism evaluation and policy analysis. To address this issue, this section introduces an exploitability-based validity assessment method. By explicitly constructing, for each agent, an approximate best response to a given strategy profile and quantifying the profit improvement induced by unilateral deviation, we assess whether the outcomes of the multi-agent simulation are close to a Nash equilibrium.

\noindent
\textbf{(1) Definition of strategy profile and payoff}

Consider a day-ahead uniform-price clearing market with $N$ generators and a time horizon of $T=96$ periods. Let $\pi_i$ denote the policy of generator $i$, and let the strategy profile be denoted by $\pi=(\pi_1,\ldots,\pi_N)$, with $\pi_{-i}$ representing the strategy profile of all agents except agent $i$. At each period, every agent submits a ten-segment stepwise bid. The Independent System Operator (ISO) clears the market by ranking bids in ascending order of price, and the marginal accepted bid determines the uniform clearing price $\lambda_t$.

Given a strategy profile $\pi$, the daily payoff (expected profit) of agent $i$ is defined as

\begin{equation}
V_i(\pi)
=
\mathbb{E}\!\left[
\sum_{t=1}^{T}
\bigl(
\lambda_t\,p_{i,t}-C_i^T(p_{i,t})
\bigr)
\right],
\qquad
i=1,\ldots,N
\label{eq:payoff_profile}
\end{equation}

where $p_{i,t}$ denotes the cleared generation output at period $t$, and $C_i^T(\cdot)$ is the total cost function of generator $i$.

\noindent
\textbf{(2) Exploitability}

In a general-sum game, the central criterion for a Nash equilibrium is that, when the strategies of all other agents are fixed, no agent should be able to substantially improve its expected payoff through unilateral deviation. To convert this criterion into a quantifiable validity metric, this section introduces the definitions of best response and exploitability.

Given $\pi_{-i}$, the best response of agent $i$ is defined as

\begin{equation}
BR_i(\pi_{-i})
\in
\arg\max_{\pi_i}
V_i(\pi_i,\pi_{-i})
\label{eq:best_response}
\end{equation}

Relative to the original policy $\pi_i$, the payoff improvement achieved by unilateral deviation to the best response is

\begin{equation}
\Delta_i(\pi)
\triangleq
V_i\bigl(BR_i(\pi_{-i}),\pi_{-i}\bigr)
-
V_i(\pi_i,\pi_{-i})
\label{eq:deviation_gain}
\end{equation}

If any agent can still significantly improve its own payoff through unilateral deviation while the other agents' policies are fixed, then the current strategy profile remains exploitable. Accordingly, the exploitability of a strategy profile is defined as

\begin{equation}
\mathcal{E}(\pi)
\triangleq
\max_i \Delta_i(\pi)
\label{eq:exploitability}
\end{equation}

When $\mathcal{E}(\pi)$ is sufficiently small, the simulation outcome can be regarded as an $\varepsilon$-Nash equilibrium.

\noindent
\textbf{(3) Exploitability based on RL approximate best responses}

In the electricity market RL-ABS setting with high-dimensional continuous states and segmented bidding action spaces, the analytical solution to Eq.~\eqref{eq:best_response} is often intractable. Therefore, this study employs an RL-based best-response search to obtain an approximate best response, which is then used to estimate $\widehat{\Delta}_i(\pi)$ and $\widehat{\mathcal{E}}(\pi)$. The core idea is to ``freeze'' the multi-agent environment into a single-agent decision problem for agent $i$, and then retrain agent $i$ to maximize its own payoff.

The specific procedure is as follows, carried out independently for each agent $i$:

\begin{enumerate}[label=\roman*.]
    \item \textbf{Freezing the opponents' policies:}
    Fix the policies of all other agents at the trained policy profile $\pi_{-i}$, and turn off exploration noise during evaluation so that their behavior remains consistent throughout the simulation. Under this setting, the environment faced by agent $i$ is transformed into a single-agent deep reinforcement learning decision process.
    
    \item \textbf{Training the approximate best response:}
    Taking $V_i(\cdot,\pi_{-i})$ as the optimization objective, agent $i$ is treated as the sole learner. Using the same RL algorithm and action representation as in the main experiment, agent $i$ is retrained under the frozen-opponent setting to obtain a policy $\widehat{BR}_i(\pi_{-i})$. To reduce the risk of local optima, training may be repeated with multiple random initializations or multiple random seeds, and the policy with the highest evaluation payoff is selected as $\widehat{BR}_i$.
    
    \item \textbf{Consistency evaluation:}
    Using the same batch of demand samples (or the same sequence of random seeds), evaluate both the baseline strategy profile $\pi$ and the deviated strategy profile $(\widehat{BR}_i(\pi_{-i}),\pi_{-i})$, yielding $\widehat{V}_i(\pi)$ and $\widehat{V}_i(\widehat{BR}_i(\pi_{-i}),\pi_{-i})$, respectively. The estimated unilateral deviation gain is then given by
    
    \begin{equation}
    \widehat{\Delta}_i(\pi)
    =
    \widehat{V}_i\bigl(\widehat{BR}_i(\pi_{-i}),\pi_{-i}\bigr)
    -
    \widehat{V}_i(\pi_i,\pi_{-i}),
    \qquad
    \widehat{\mathcal{E}}(\pi)=\max_i \widehat{\Delta}_i(\pi)
    \label{eq:estimated_exploitability_en}
    \end{equation}

    If the resulting $\widehat{\mathcal{E}}(\pi)$ is sufficiently small, the simulation outcome may be regarded as an approximate $\varepsilon$-Nash equilibrium. Conversely, if $\widehat{\mathcal{E}}(\pi)$ is evidently large, this indicates that the current strategy profile still admits room for improvement through unilateral deviation, and thus cannot be regarded as an approximate $\varepsilon$-Nash equilibrium.

\end{enumerate}

\section{Validity Assessment of the DPMP}

\subsection{Single-Agent Optimality-Gap Assessment of DPMP}

\subsubsection{Research Questions and Experimental Design}

To assess the effectiveness of DPMP in single-agent electricity market bidding tasks, this study constructs a benchmark environment in which the theoretical optimal profit $\Pi_t^*$ can be computed explicitly. In this environment, the opponents' stepwise bids, the market-clearing rules, the reward definition, and the state transition mechanism are all kept fixed, while the optimality gap is adopted as the unified evaluation criterion. Guided by this objective, this subsection focuses on the following two research questions:

RQ1: Under a fixed reinforcement learning algorithm, does the post-processing mapping systematically affect the learnability of the policy and its ability to approach the optimum, and does DPMP exhibit a clear advantage? To answer this question, DPMP is compared with three engineering baselines based on commonly used post-processing mappings, namely SORT, CLIP, and PROJECT. The core purpose of this comparison is to isolate the effect of the reinforcement learning algorithm and determine whether DPMP can deliver a smaller optimality gap and higher sample efficiency.

RQ2: When DPMP is fixed, how do different reinforcement learning algorithms differ in their ability to approach the optimal policy and in training stability, and is DPMP compatible with mainstream reinforcement learning algorithms? To address this question, this subsection compares the performance of A2C, TRPO, PPO, and DDPG under the unified DPMP representation.

To answer the above two research questions, this subsection adopts a controlled two-axis comparative design. The “two axes” refer to two experimental factor axes constructed to correspond to the two research questions, with only one core factor varied along each axis while all other conditions are held constant. The horizontal experiment (Axis A) is designed to identify the effect of the post-processing mapping: the learning algorithm is fixed as PPO, and only the mapping from actions to bid curves is changed, with DPMP, SORT, PROJECT, and CLIP considered as four alternative implementations. Since the algorithm, environment, and training budget are all kept unchanged, performance differences along this axis can be attributed primarily to the post-processing mapping itself. The vertical experiment (Axis B) is designed to identify the effect of the reinforcement learning algorithm: the bid parameterization is fixed as DPMP, and only the optimization algorithm is varied, with A2C, TRPO, PPO, and DDPG compared. Under this setting, the observed differences can be attributed mainly to variations in the search and optimization capabilities of different algorithms within the same feasible bid representation space.

To ensure that the results obtained along the two identification axes are comparable under a common standard, all experiments in this study adopt the same evaluation metric system. First, the average daily optimality gap is used to measure the average proximity of a method to the theoretical optimal profit over the entire scheduling horizon. Second, the number of environment interaction steps required to reach a predefined optimality threshold is used to characterize sample efficiency. Finally, the magnitude of fluctuations during training is used to characterize training stability. Based on this unified evaluation framework, Table~\ref{tab:exp_grid1} summarizes the settings of each experimental group and their corresponding identification objectives.

\begin{table}[H]
\centering
\caption{Experiment grid}
\label{tab:exp_grid1}
\small
\renewcommand{\arraystretch}{1.15}
\setlength{\tabcolsep}{4pt}

\begin{tabularx}{\linewidth}{
C{1.15cm}
C{1.55cm}
Y
C{1.55cm}
C{2.05cm}
C{2.85cm}
}
\toprule
\makecell[c]{Group\\ID} &
\makecell[c]{Research\\Axis} &
Objective &
\makecell[c]{RL\\Algorithm} &
\makecell[c]{Type of\\Post-Processing\\Mapping} &
\makecell[c]{Market/Opponent\\Setting} \\
\midrule

A1 & A &
\makecell[c]{Assess the optimality\\gap and training\\stability of DPMP} &
PPO & DPMP &
\makecell[c]{Section 6.1.2\\market setting:\\fixed opponents'\\stepwise bids} \\

A2 & A &
\makecell[c]{Baseline 1: SORT-\\based post-processing\\for monotonicity\\enforcement; compared\\with DPMP} &
PPO & SORT &
\makecell[c]{Same as\\above} \\

A3 & A &
\makecell[c]{Baseline 2: PROJECT-\\based post-processing;\\compared with DPMP} &
PPO & PROJECT &
\makecell[c]{Same as\\above} \\

A4 & A &
\makecell[c]{Baseline 3: CLIP-\\based post-processing;\\compared with DPMP} &
PPO & CLIP &
\makecell[c]{Same as\\above} \\

B1 & B &
\makecell[c]{Performance of DPMP\\under A2C} &
A2C & DPMP &
\makecell[c]{Same as\\above} \\

B2 & B &
\makecell[c]{Performance of DPMP\\under TRPO} &
TRPO & DPMP &
\makecell[c]{Same as\\above} \\

B3 & B &
\makecell[c]{Performance of DPMP\\under PPO} &
PPO & DPMP &
\makecell[c]{Same as\\above} \\

B4 & B &
\makecell[c]{Performance of DPMP\\under DDPG} &
DDPG & DPMP &
\makecell[c]{Same as\\above} \\
\bottomrule
\end{tabularx}
\end{table}

To ensure that all experiments are compared under a common standard, the following evaluation metrics are defined:

\begin{enumerate}[label=\roman*.]
    \item Average Daily Optimality Gap
    
    \begin{equation}
    \bar{\delta}
    =
    \frac{1}{T}
    \sum_{t=1}^{T}
    \frac{\Pi_t^*-\Pi_t}{\left|\Pi_t^*\right|}
    \label{eq:avg_daily_opt_gap}
    \end{equation}
    
    \item Sample Efficiency: the number of environment interaction steps required to reach the threshold $\bar{\delta}\le 0.1$.
    
    \item Training Stability: the variance of the optimality gap.
\end{enumerate}

\subsubsection{Environment Setup}

To obtain a computable theoretical optimal strategy and thereby quantify the optimality gap, this study constructs a multi-period electricity market environment involving a single generating unit. In this environment, the opponent’s bids are fixed and demand exhibits structured fluctuations, thus providing support for the subsequent measurement of the optimality gap.

\noindent
\textbf{(1) Market setup}

The market is modeled as a sequential decision-making process of length $T=96$. Each episode corresponds to a within-day clearing sequence. At each period $t\in\{0,\ldots,T-1\}$, the Agent and the opponent each submit a 10-segment stepwise bid curve. The price cap is set to $1000$ CNY/MWh. The market clears against the prevailing demand using a uniform clearing price and returns the resulting profit as the reward signal.

\noindent
\textbf{(2) Demand setup}

The demand at period $t$, denoted by $D_t$, is composed of an intraday periodic component and a Gaussian perturbation, and is clipped to the feasible interval:

\begin{equation}
D_t
=
\operatorname{clip}
\!\left(
500+300\sin\!\left(2\pi t/T-\pi/2\right)+\varepsilon_t,\;0,\;1000
\right),
\qquad
\varepsilon_t\sim\mathcal{N}(0,25^2)
\label{eq:demand_process}
\end{equation}

This setting ensures that demand exhibits a pronounced peak--valley pattern within the day, which facilitates the observation of policy adaptability across different periods. Meanwhile, the noise term introduces a certain degree of randomness, preventing the learning process from degenerating into deterministic fitting.

\noindent
\textbf{(3) Opponent bid setup}

The Opponent is assigned a fixed 10-segment stepwise bid, with each segment having a width of 100 MW and a total offered quantity of 1000 MW. The segment prices are fixed at $[20,25,30,35,40,45,50,55,60,65]$.

The rationale for fixing the opponent is that the market-clearing price can only take values from a finite set $\{p_j\}$, which lays the foundation for constructing the theoretical optimal profit in Section 6.1.3.

\noindent
\textbf{(4) Agent cost modeling}

The marginal cost curve of the generating unit, $MC(q)$, is discretized over $q\in[0,Q_{\max}]$ using $101$ points. It is specified in the following power-function form:

\begin{equation}
MC(q_i)
=
a+b\left(\frac{q_i}{Q_{\max}}\right)^{\gamma},
\qquad
\gamma\sim \mathrm{Uniform}(1,2)
\label{eq:agent_mc}
\end{equation}

where $a=20$ and $b=300$, and $q_i$ denotes equally spaced grid points. The total cost $C(q)$ is then obtained through discrete accumulation using trapezoidal integration:

\begin{equation}
C(q_i)
=
\sum_{k=1}^{i}
\frac{MC(q_k)+MC(q_{k-1})}{2}\,(q_k-q_{k-1})
\label{eq:agent_cost_trap}
\end{equation}

This construction has two advantages:

\begin{enumerate}[label=\roman*.]
    \item the cost curve is strictly increasing and convex, which is consistent with the typical characteristics of the marginal cost of a single generating unit;
    \item $C(q)$ can be computed numerically with high precision, thereby providing a basis for calculating the theoretical optimal profit.
\end{enumerate}

\subsubsection{Calculation of the Theoretical Optimal Profit}

In the optimality-gap assessment of reinforcement learning algorithms, the core task is to construct a verifiable upper bound on optimality, which can serve as a benchmark for alternative bidding strategies. This section details how the theoretical optimal profit $\Pi_t^*$ is computed on a period-by-period basis according to the environment settings and the market mechanism, so as to evaluate the learning performance of the agent.

For fixed 10-segment stepwise bids on the supply side, there are 10 discrete price levels $\{p_j\}_{j=0}^{9}$, ordered from low to high. Under this structure, if the agent attains the optimal profit, the market-clearing price can only lie at or near these discrete levels. Therefore, the candidate clearing-price set is restricted to $\{p_j\}$, thereby reducing the theoretical optimum to an enumeration problem over a finite set. The key is to characterize the feasible output of the unit at a given offer price $p$. If the unit's marginal cost curve is monotonically increasing $MC(q)$, we define

\begin{equation}
q^{MC}(p)
=
\sup\left\{
q\in[0,Q_{\max}] \,:\, MC(q)\le p
\right\}
\label{eq:q_mc_p}
\end{equation}

which represents the maximum output that does not incur a loss at price $p$. At a candidate price level $p_j$, the total output available from lower price segments is $j\Delta q$. Therefore, when the price increases to $p_j$, if the following feasibility condition is satisfied,

\begin{equation}
j\Delta q+q^{MC}(p_j)\ge D_t
\label{eq:feasibility_pj}
\end{equation}

then this price level is capable of covering the demand $D_t$. The residual demand that must be supplied by the unit is then

\begin{equation}
q_j(D_t)=D_t-j\Delta q
\label{eq:residual_qj}
\end{equation}

Let the unit's total cost function be $C(q)=\int_0^q MC(x)\,dx$ (estimated via discrete integration/interpolation). Then, the candidate profit at price level $p_j$ is

\begin{equation}
\Pi_t(p_j)=p_j\cdot q_j(D_t)-C\bigl(q_j(D_t)\bigr)
\label{eq:candidate_profit_pj}
\end{equation}

Finally, the theoretical optimal profit at period $t$ is defined as

\begin{equation}
\Pi_t^*
=
\max_{j\in\{0,\ldots,9\}} \Pi_t(p_j)
\quad
\text{s.t.}\quad
q_j(D_t)>0,\;
j\Delta q + q^{MC}(p_j)\ge D_t
\label{eq:theoretical_opt_profit}
\end{equation}

The corresponding upper-bound realization $(p_t^*,q_t^*)$ is obtained simultaneously.The implementation is given in Algorithm~\ref{alg:theoretical_opt_profit_en}.

\begin{algorithm}[H]
\caption{Theoretical Optimal Profit (Single Period)}
\label{alg:theoretical_opt_profit_en}
\small
\textbf{Input:} demand $D_t$, price levels $\{p_j\}_{j=0}^{9}$, segment output $\Delta q$, unit $MC(\cdot)$ and $C(\cdot)$

\textbf{Output:} $\Pi_t^*$ and the corresponding $(p_t^*, q_t^*)$

\begin{algorithmic}[1]
\State Initialize $\Pi_t^* \gets -\infty$
\For{$j=0,\ldots,9$}
    \State Compute $q^{MC}(p_j)$
    \If{$j\Delta q + q^{MC}(p_j) < D_t$}
        \State \textbf{continue}
    \EndIf
    \State $q \gets D_t - j\Delta q$
    \If{$q \le 0$}
        \State \textbf{continue}
    \EndIf
    \State $\Pi \gets p_j q - C(q)$
    \If{$\Pi > \Pi_t^*$}
        \State $\Pi_t^* \gets \Pi$
        \State $(p_t^*, q_t^*) \gets (p_j, q)$
    \EndIf
\EndFor
\State \Return $\Pi_t^*$ and $(p_t^*, q_t^*)$
\end{algorithmic}
\end{algorithm}

\subsubsection{Implementation of DPMP and the SORT/CLIP/PROJECT Baselines}
To ensure a controlled horizontal comparison, only the post-processing mapping from the policy-network output $u$ to the bid curve is varied, while the reinforcement learning algorithm (PPO) and the market-clearing structure are kept fixed. The four modelling approaches are unified under the same interface:

\begin{equation}
(q^{\mathrm{edge}},\,p)=h_{\mathrm{mode}}(u),
\qquad
u\in[0,1]^{2K},\quad K=10
\label{eq:unified_interface}
\end{equation}

where $q^{\mathrm{edge}}=(q_0,\ldots,q_K)$ denotes the generation output breakpoints ($q_0=0,\ q_K=Q^{\max}$), and $p=(p_1,\ldots,p_K)$ denotes the segment prices. These variables satisfy the common constraints in electricity markets: the generation output breakpoints must be strictly increasing, and the segment prices must be nondecreasing and bounded, i.e., $p_i\in[0,\bar{p}]$. The implementation of DPMP has already been detailed in Section~5. We next describe the implementations of the SORT/CLIP/PROJECT baselines.

\noindent
\textbf{(1) SORT baseline}

SORT is a common engineering practice. The generation output breakpoints are fixed as 10 equally spaced segments:

\begin{equation}
q_i=i\cdot \frac{Q_{\max}}{K},
\qquad
i=0,\ldots,K
\label{eq:sort_q_fixed}
\end{equation}

The prices are first linearly mapped to $[0,\bar{p}]$, and then globally sorted to enforce monotonicity:

\begin{equation}
\tilde{p}_i=\bar{p}\,u_i^{(p)},
\qquad
p=\operatorname{sort}(\tilde{p})
\label{eq:sort_price_mapping}
\end{equation}

The advantage of this method is its extreme simplicity while guaranteeing monotonicity. However, mechanistically, it is a many-to-one mapping: a large number of different $\tilde{p}$ values are mapped to the same $p$. Therefore, it does not satisfy the injectivity and invertibility requirements imposed on the post-processing mapping under the policy gradient framework, thereby leading to objective mismatch (a rigorous analysis is provided in Appendix A).

\noindent
\textbf{(2) CLIP baseline}

Like SORT, CLIP also fixes the generation output breakpoints to be equally spaced. However, instead of sorting the prices, it adopts a recursively defined minimal-modification projection:

\begin{equation}
\tilde{p}_i=\bar{p}\,u_i^{(p)},
\qquad
p_1=\tilde{p}_1,
\qquad
p_i=\max(p_{i-1},\tilde{p}_i)\quad (i\ge 2)
\label{eq:clip_price_mapping}
\end{equation}

That is, $p=\operatorname{cummax}(\tilde{p})$. This method guarantees monotonicity and compliance with the price cap. However, the above recursion compresses a continuous region of the input space onto the equivalence constraint faces $\{p_i=p_{i-1}\}$, causing discontinuous density accumulation of the executed-action distribution on these faces and thereby violating the continuity conditions required for policy gradient implementation.

\noindent
\textbf{(3) PROJECT baseline}

The PROJECT baseline projects the raw price sequence output by the network onto a feasible set satisfying the monotonicity and boundary constraints, thereby ensuring a feasible bid curve with the minimum perturbation in the $L_2$ sense.

With the generation output divided into $K$ equally spaced segments, the policy network outputs a raw $K$-dimensional vector $u^{(p)}\in\mathbb{R}^K$. This vector is first mapped to a raw price sequence $x\in[0,\bar{p}]^K$. Let $\mathcal{C}$ denote the convex feasible set of bounded nondecreasing sequences:

\begin{equation}
\mathcal{C}
=
\left\{
p\in\mathbb{R}^K
\ \middle|\
0\le p_1\le p_2\le \cdots \le p_K\le \bar{p}
\right\}
\label{eq:project_feasible_set}
\end{equation}

The core of PROJECT is the Euclidean projection of $x$ onto $\mathcal{C}$, i.e., the nearest-point projection in the least-squares sense:

\begin{equation}
p
=
\Pi_{\mathcal{C}}(x)
=
\arg\min_{y\in\mathcal{C}}\|y-x\|_2^2
=
\arg\min_{y\in\mathcal{C}}\sum_{i=1}^{K}(y_i-x_i)^2
\label{eq:project_projection}
\end{equation}

The resulting $p=(p_1,\ldots,p_K)$ naturally satisfies $0\le p_1\le \cdots \le p_K\le \bar{p}$. In this study, the projection is efficiently implemented using the PAV (Pool-Adjacent-Violators) algorithm \cite{de2010isotone}. Based on the projected price sequence, the feasible stepwise bid curve is constructed as

\begin{equation}
\phi(q)=p_i,
\qquad
q\in(q_{i-1},q_i],\quad i=1,\ldots,K
\label{eq:project_step_function}
\end{equation}

thus yielding the final bid steps used for market clearing, $\{((q_{i-1},q_i],p_i)\}_{i=1}^{K}$.

\begin{algorithm}[H]
\caption{Post-processing mapping algorithm $h_{\mathrm{mode}}(u)$}
\label{alg:postprocess_mapping}
\small
\begin{algorithmic}[1]

\Statex \textbf{Input:}
\Statex \hspace{\algorithmicindent}$u=(u^{(q)},u^{(p)})\in(0,1)^{2K}$
\Statex \hspace{\algorithmicindent}$Q^{\max}$: maximum generation output; $\bar{p}$: price cap
\Statex \hspace{\algorithmicindent}$\mathrm{mode}\in\{\mathrm{DPMP},\mathrm{SORT},\mathrm{CLIP},\mathrm{PROJECT}\}$
\Statex \hspace{\algorithmicindent}Hyperparameters: $k,\alpha,\varepsilon$ ($\varepsilon\approx 10^{-6}$, used only for numerical stability)

\Statex \textbf{Output:}
\Statex \hspace{\algorithmicindent}Generation output breakpoints $q^{\mathrm{edge}}=(q_0,q_1,\ldots,q_K)\in\mathbb{R}^{K+1}$, where $q_0=0,\ q_K=Q^{\max}$
\Statex \hspace{\algorithmicindent}Segment prices $p=(p_1,\ldots,p_K)\in\mathbb{R}^K$, satisfying $0\le p_1\le\cdots\le p_K\le \bar{p}$

\If{$\mathrm{mode}=\mathrm{DPMP}$}
    \State $w \gets \operatorname{softmax}(u^{(q)})$
    \For{$i=1,\ldots,K$}
        \State $\Delta q_i \gets Q^{\max} w_i$
    \EndFor
    \State $q_0 \gets 0$
    \For{$i=1,\ldots,K$}
        \State $q_i \gets q_{i-1}+\Delta q_i$
    \EndFor
    \For{$i=1,\ldots,K$}
        \State $\mathrm{raw}_i \gets k\sum_{j=1}^{i}u_j^{(p)}$
    \EndFor
    \For{$i=1,\ldots,K$}
        \State $p_i \gets \bar{p}\bigl(1-\exp(-\alpha\,\mathrm{raw}_i)\bigr)$
    \EndFor
    \State \textbf{return} $(q^{\mathrm{edge}},p)$

\ElsIf{$\mathrm{mode}=\mathrm{SORT}$}
    \For{$i=0,\ldots,K$}
        \State $q_i \gets i\cdot Q^{\max}/K$
    \EndFor
    \For{$i=1,\ldots,K$}
        \State $\tilde{p}_i \gets \bar{p}\,u_i^{(p)}$
    \EndFor
    \State $p \gets \operatorname{sort}(\tilde{p})$
    \State \textbf{return} $(q^{\mathrm{edge}},p)$

\ElsIf{$\mathrm{mode}=\mathrm{CLIP}$}
    \For{$i=0,\ldots,K$}
        \State $q_i \gets i\cdot Q^{\max}/K$
    \EndFor
    \For{$i=1,\ldots,K$}
        \State $\tilde{p}_i \gets \bar{p}\,u_i^{(p)}$
    \EndFor
    \State $p_1 \gets \tilde{p}_1$
    \For{$i=2,\ldots,K$}
        \State $p_i \gets \max(p_{i-1},\tilde{p}_i)$
    \EndFor
    \State \textbf{return} $(q^{\mathrm{edge}},p)$

\ElsIf{$\mathrm{mode}=\mathrm{PROJECT}$}
    \For{$i=0,\ldots,K$}
        \State $q_i \gets i\cdot Q^{\max}/K$
    \EndFor
    \For{$i=1,\ldots,K$}
        \State $x_i \gets \operatorname{clip}\!\bigl(\bar{p}\cdot g(u_i^{(p)}),\,0,\,\bar{p}\bigr)$
    \EndFor
    \State $p \gets \Pi_{\mathcal{C}}(x)$
    \State \textbf{return} $(q^{\mathrm{edge}},p)$
\EndIf

\end{algorithmic}
\end{algorithm}

\subsubsection{Horizontal experiments and result analysis}

To answer RQ1, this subsection conducts a comparative evaluation of the four post-processing mappings—DPMP, SORT, CLIP, and PROJECT—under the same reinforcement learning algorithm, the same training configuration, the same market environment, the same opponents’ stepwise bids, and the same reward definition. Since this subsection changes only the post-processing mapping from actions to bid curves while keeping all other conditions fixed, the observed performance differences can be attributed primarily to the post-processing mapping itself, rather than to the learning algorithm or environmental perturbations.

\noindent
\textbf{(1) Evolution of stepwise bid-curve morphology}

\begin{figure*}[!t]
\centering
\caption{Evolution of stepwise bid curves over episodes}
\label{fig:staircase_evolution}

\setlength{\tabcolsep}{0pt}
\renewcommand{\arraystretch}{0}

\begin{tabular}{|c|c|}
\hline
\includegraphics[width=0.49\textwidth]{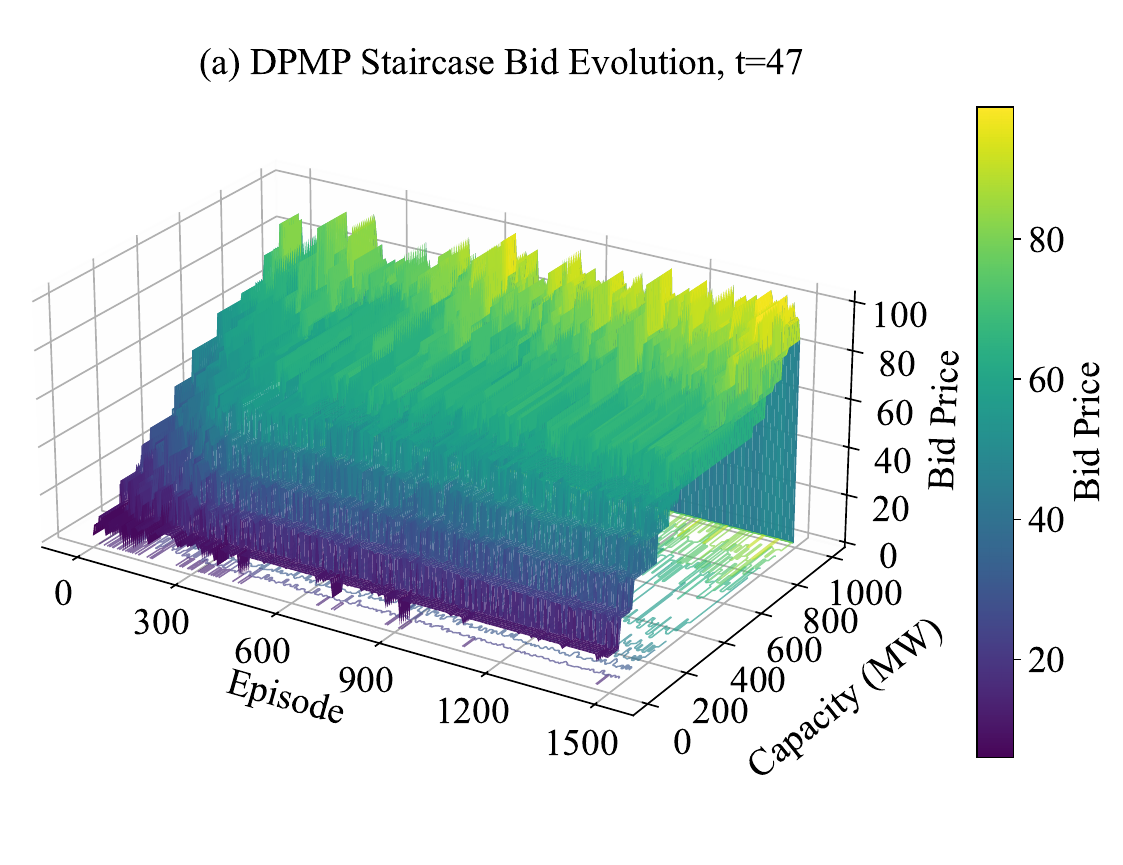} &
\includegraphics[width=0.49\textwidth]{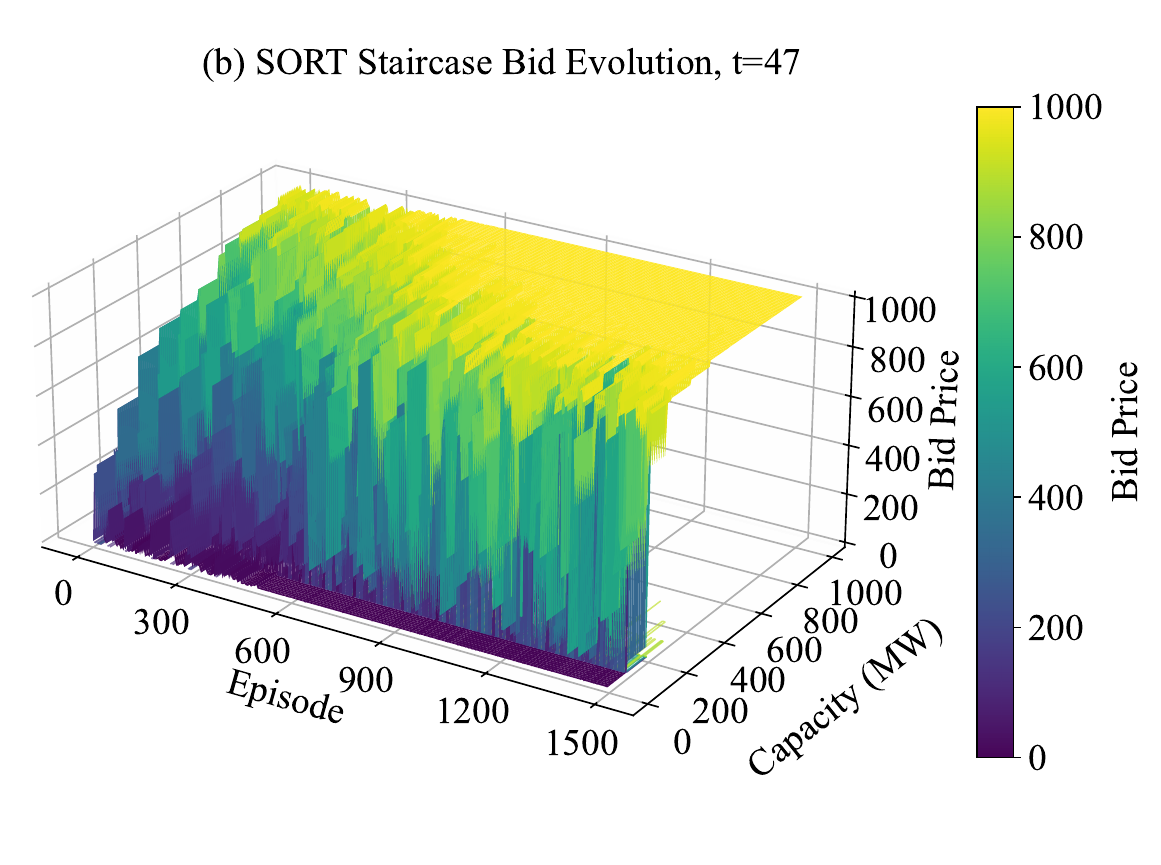} \\
\hline
\includegraphics[width=0.49\textwidth]{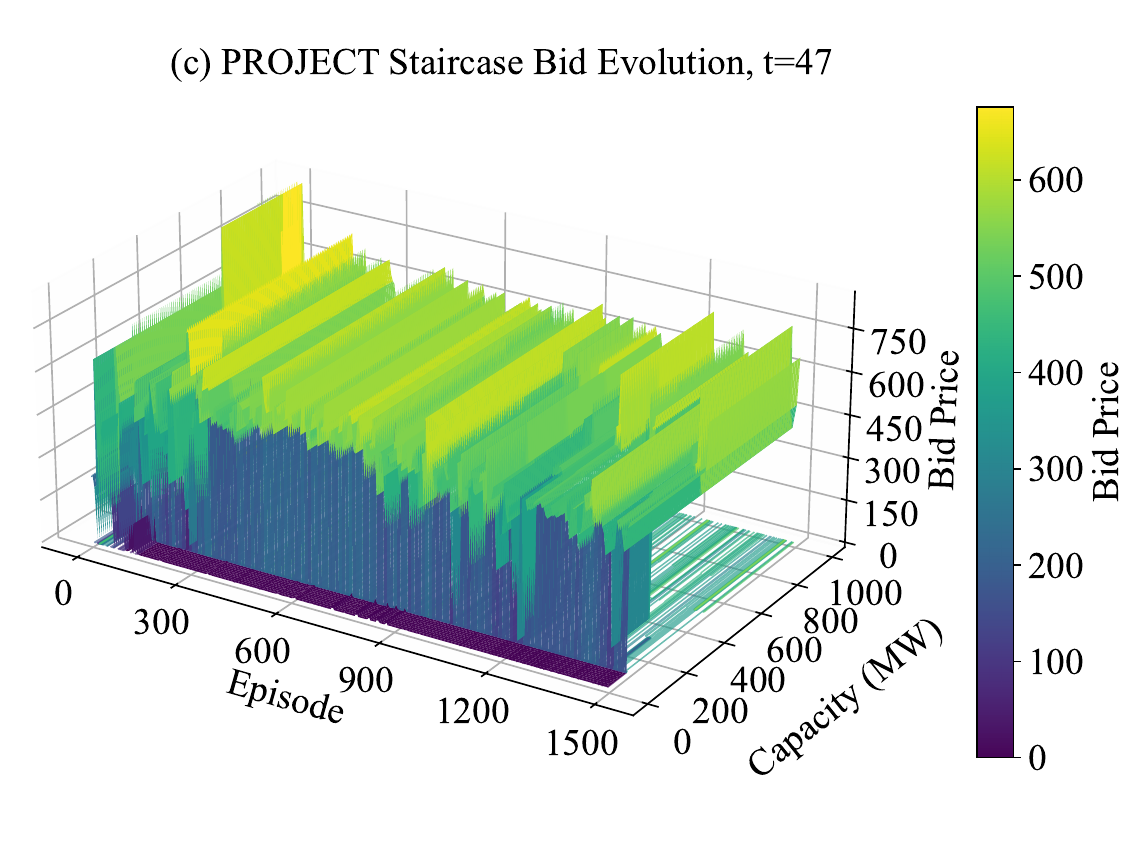} &
\includegraphics[width=0.49\textwidth]{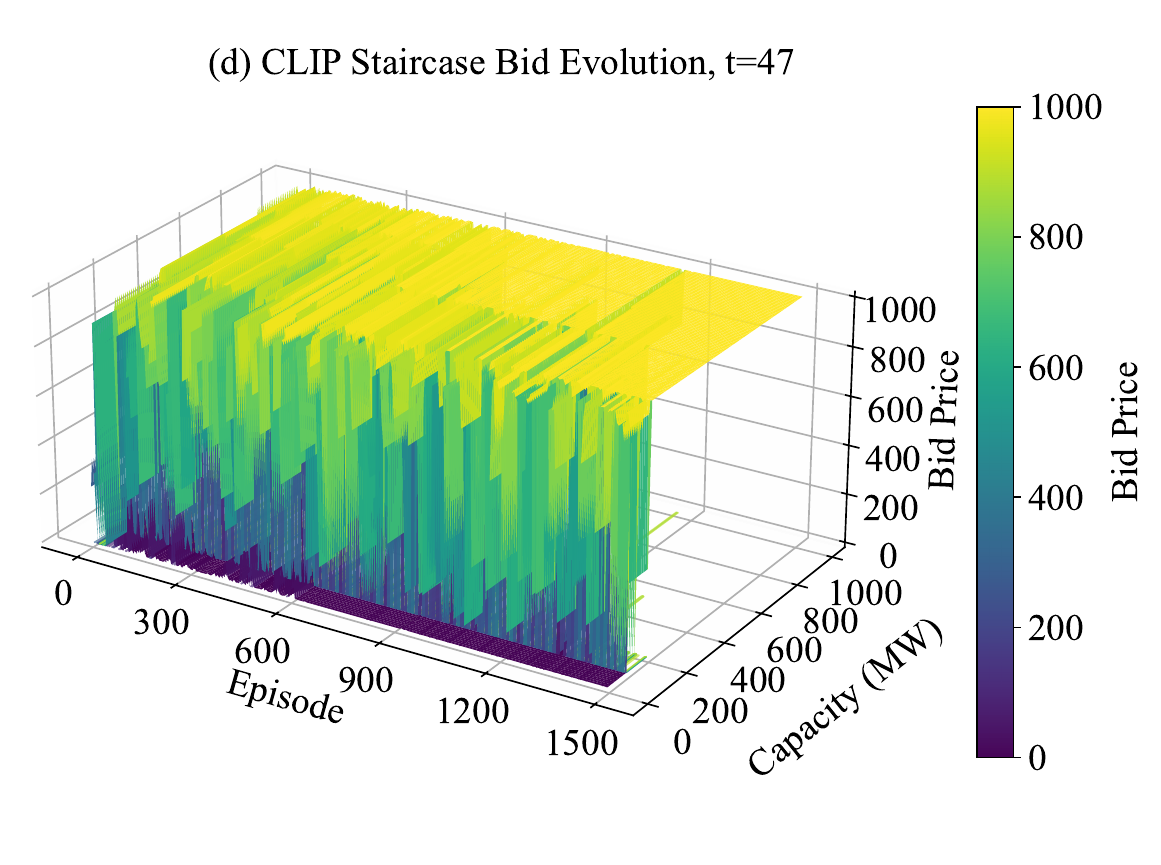} \\
\hline
\end{tabular}
\end{figure*}

Figure~\ref{fig:staircase_evolution} illustrates, for the four methods during training, the evolution of the stepwise bid curves at time $t=47$ over episodes (with episode on the x-axis, generation output on the y-axis, and bid price on the z-axis). This figure not only visualizes the training process, but also directly reveals the morphology of the policy search trajectory in the feasible bid space under different post-processing mappings. If a method forms a stable stepwise structure relatively early in training and exhibits a continuous and smooth surface evolution across episodes, this usually indicates that its action representation and post-processing mapping are more amenable to optimization. By contrast, if the surface persistently exhibits large-scale boundary adhesion, abrupt discontinuities, or high-frequency oscillations, this often suggests substantial information compression, boundary concentration, or gradient distortion during training.

Overall, DPMP gradually converges in the middle and late stages of training to a structurally stable and well-stratified stepwise surface. Its bid prices exhibit a consistent monotone structure as generation output increases, and the evolutionary trajectory across episodes remains relatively continuous, yielding an overall smooth surface with only limited local exploratory perturbations. Both SORT and CLIP exhibit pronounced boundary concentration. As can be seen from the figure, the bid surfaces rapidly approach the price cap over large ranges of generation output and episodes, while the low-generation-output region frequently adheres to the lower boundary, forming a clear two-layer structure concentrated near the upper and lower bounds, with a relatively thin effective transition layer in between. This pattern indicates that, after post-processing, the policy outputs are compressed toward the boundary of the feasible set, thereby weakening the distinguishable structure in the interior. The boundary concentration in PROJECT is somewhat alleviated, but its surface still exhibits noticeable banded jumps and local spikes. Across episodes, the surface shows more frequent segment-wise discontinuities and plateau reconfigurations, indicating that although projection-based feasibility restoration can ensure constraint satisfaction, structural instability during training still persists.

A closer inspection of the three-dimensional evolution of DPMP further shows that bid prices in the low-generation-output region stabilize relatively early across episodes, whereas those in the high-generation-output region still retain a certain degree of fluctuation in the later stages of training. This phenomenon does not imply that DPMP fails to converge in the high-generation-output region; rather, it reflects differences in reward sensitivity across generation-output intervals under the market-clearing mechanism. Specifically, under fixed opponent bids and a given demand condition, the segments that truly affect the market-clearing outcome and profit are usually those located near the marginal clearing position. For these low- and medium-generation-output segments, which are highly sensitive to clearing probability and transaction price, an inappropriate bid is immediately reflected in profit variation. Reinforcement learning can therefore obtain stronger and more stable feedback signals, driving faster parameter convergence in these segments. By contrast, some high-generation-output segments remain non-marginal in most episodes, meaning that even if their bid prices are adjusted, the actual market-clearing outcome does not change. Their influence on the immediate reward is therefore weaker, so the corresponding parameters receive effective learning signals that are sparser and noisier. As a result, the surface appears more stable across episodes in the low-generation-output region, while a certain degree of exploratory fluctuation is still retained in the high-generation-output region. It is important to emphasize that, even under such conditions, DPMP still preserves the overall monotonicity of the stepwise structure and the continuity of the surface. This indicates that its parameterization is able to strike a favorable balance between continued local exploration and global structural stability: training resources are preferentially allocated to the critical generation-output intervals that genuinely affect market clearing and profit, rather than being wasted on indiscriminate boundary oscillations over the entire generation-output range.

\noindent
\textbf{(2) Comparison of profit curves and optimality-gap curves}

\begin{figure}[H]
    \centering
    \includegraphics[width=0.55\linewidth]{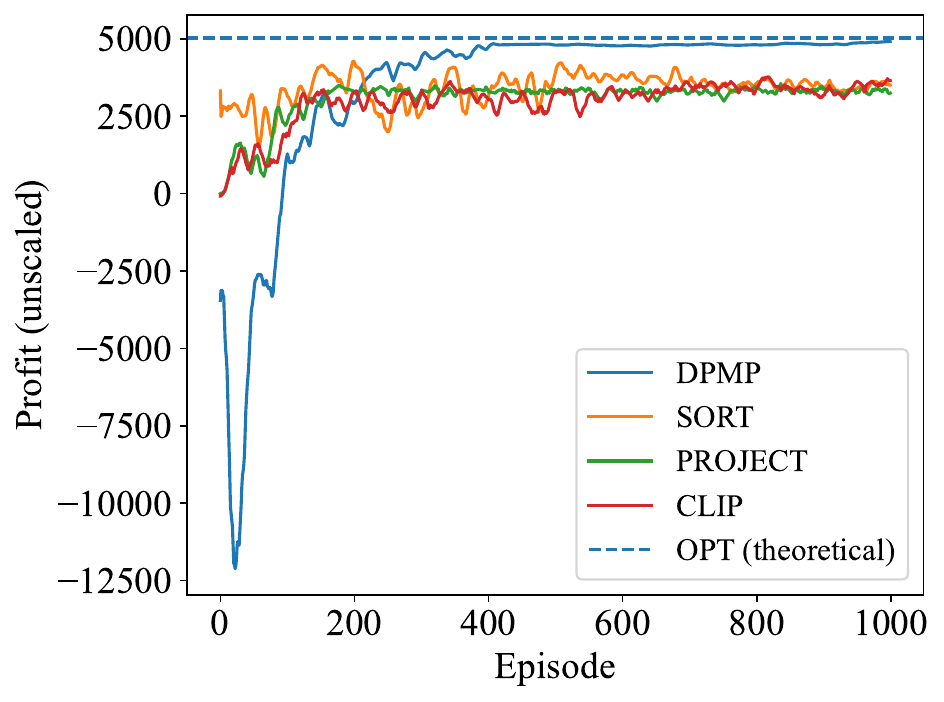}
    \caption{Profit curves of the four methods (DPMP/SORT/PROJECT/CLIP).}
    \label{fig:profit_curves}
\end{figure}

Figure~\ref{fig:profit_curves} and Figure~\ref{fig:four_algorithms_relative_opt_gap} present, respectively, the average profit per episode and the optimality-gap curves of the four methods during training, while Table~\ref{tab:horizontal_opt_gap_stats} reports the corresponding statistics of the optimality gap. The results exhibit two distinctly different learning trajectories: DPMP follows a ``high exploration cost--rapid transition--high-level stabilization'' pattern, whereas SORT, PROJECT, and CLIP exhibit an ``early rapid profit increase--mid-to-late-stage suboptimal stagnation'' pattern. This divergence is consistent with the qualitative findings from the morphological analysis of the stepwise bid curves in the previous subsection.

In terms of steady-state performance (over the last 10\% of episodes), the steady-state relative optimality gap of DPMP is only 3.26\%$\pm$0.73\%, which is substantially lower than that of SORT (30.95\%$\pm$2.42\%), PROJECT (33.12\%$\pm$3.31\%), and CLIP (31.62\%$\pm$3.50\%). In terms of steady-state mean values, the relative optimality gap of DPMP is lower than those of SORT, PROJECT, and CLIP by 27.69, 29.86, and 28.36 percentage points, respectively, corresponding to relative reductions of 89.47\%, 90.16\%, and 89.69\%. This further implies that, although the three baseline methods can achieve stable positive profits, their policies remain trapped in a suboptimal region that is approximately 30\% away from the theoretical optimum over the long run. By contrast, DPMP has stably entered a near-optimal convergence regime.

In terms of sample efficiency, using a 10\% optimality gap as the threshold, only DPMP reaches the target within the training horizon: it attains the 10\% optimality-gap threshold at episode 328 and further reaches the 5\% threshold at episode 398. By contrast, SORT, PROJECT, and CLIP all fail to reach the 10\% threshold within 1000 episodes.

Overall, the profit curves, optimality-gap curves, and the associated statistics consistently show that the core advantage of DPMP does not lie in achieving higher profits at the early stage of training, but in delivering higher steady-state profits, stronger approximation to the theoretical optimum, and better late-stage stability at an acceptable exploration cost.

\begin{figure}[H]
    \centering
    \includegraphics[width=0.65\linewidth]{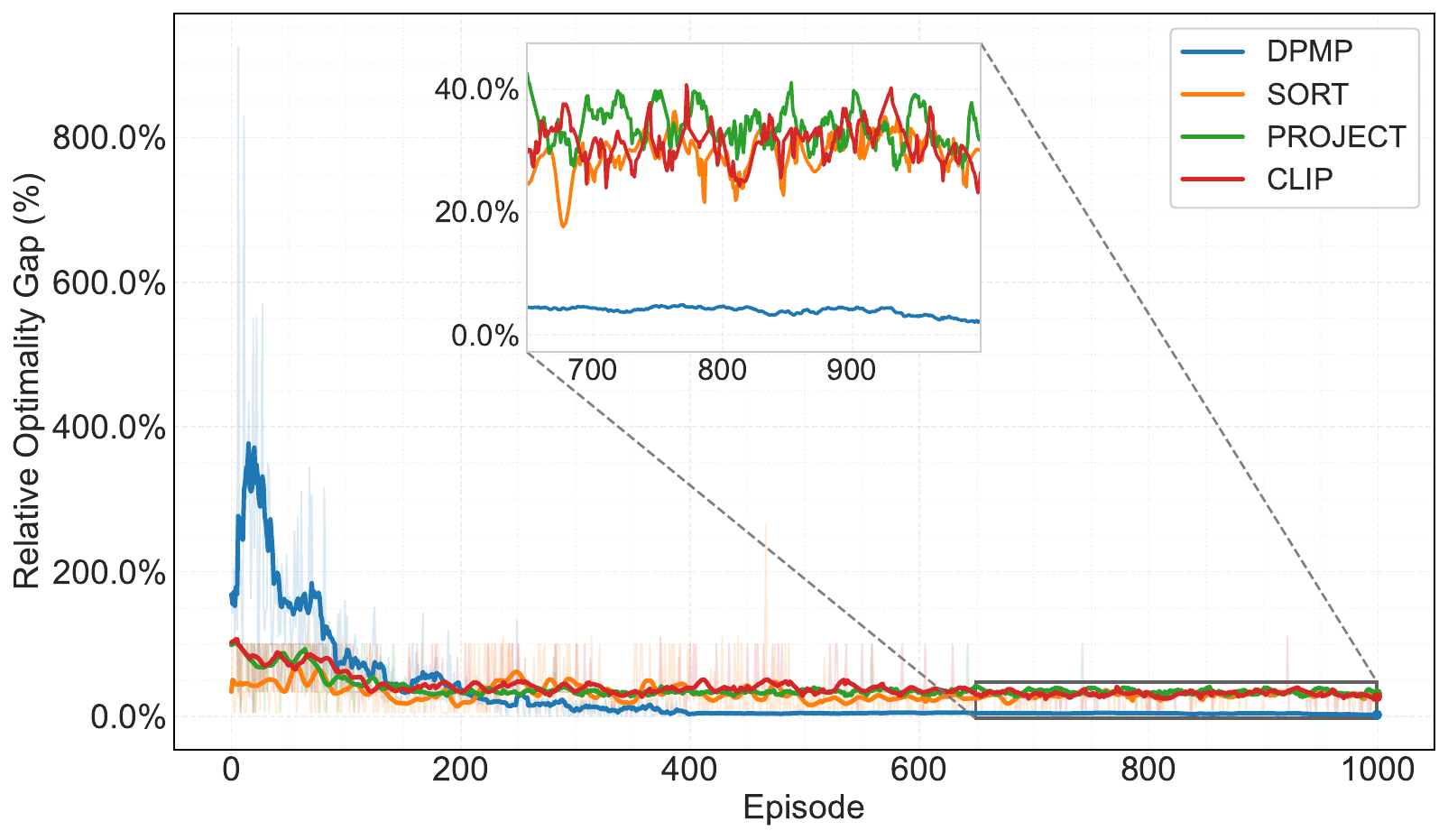}
    \caption{Optimality-gap curves of the four methods (DPMP/SORT/PROJECT/CLIP).}
    \label{fig:four_algorithms_relative_opt_gap}
\end{figure}

\begin{table}[H]
\centering
\caption{Statistics of optimality-gap curves in the horizontal experiments}
\label{tab:horizontal_opt_gap_stats}
\small
\renewcommand{\arraystretch}{1.15}
\setlength{\tabcolsep}{4pt}

\begin{tabularx}{\linewidth}{
C{1.3cm}
C{2.7cm}
C{2.0cm}
C{2.0cm}
C{2.3cm}
C{1.9cm}
}
\toprule
Method &
\makecell[c]{Final steady-state\\optimality gap\\(last 10\%,\\mean$\pm$std)} &
\makecell[c]{Episode reaching\\the 10\% gap} &
\makecell[c]{Episode reaching\\the 5\% gap} &
\makecell[c]{Best gap\\(minimum MA\\over all episodes)} &
\makecell[c]{Compliance rate in the\\last 10\%\\($\le$10\%)} \\
\midrule
DPMP    & $3.26\%\pm0.73\%$  & 328 & 398 & $2.04\%$  & $100.0\%$ \\
SORT    & $30.95\%\pm2.42\%$ & --  & --  & $14.82\%$ & $0.0\%$   \\
PROJECT & $33.12\%\pm3.31\%$ & --  & --  & $26.83\%$ & $0.0\%$   \\
CLIP    & $31.62\%\pm3.50\%$ & --  & --  & $23.55\%$ & $0.0\%$   \\
\bottomrule
\end{tabularx}
\end{table}

\subsubsection{Vertical experiments and analysis of results}

To answer RQ2, this subsection continues to adopt a controlled-variable design: the bid parameterization is fixed as DPMP, the market environment and the opponents’ stepwise bids are kept unchanged, and the reward definition and training configuration are held constant, while only the reinforcement learning algorithm is varied (A2C, TRPO, PPO, and DDPG). Therefore, the observed performance differences can be attributed primarily to the algorithms themselves, rather than to differences in action representation or environmental perturbations.

Figure~\ref{fig:four_algorithms_relative_opt_gap} presents the relative optimality-gap curves of the four algorithms under DPMP, and Table~\ref{tab:vertical_opt_gap_stats} reports the corresponding statistics. Overall, all four algorithms exhibit a convergent pattern characterized by a rapid decline followed by entry into a steady-state plateau. Moreover, the pass rates in the last 10\% of episodes ($\delta \le 10\%$) all reach 100\%, indicating that DPMP does not rely on any particular learning algorithm, but can instead provide a consistent and optimizable monotone feasible bid representation for mainstream reinforcement learning methods.

In terms of steady-state performance (last 10\% of episodes), the four algorithms can be divided into two categories: PPO and DDPG exhibit a low-gap steady-state pattern, whereas A2C and TRPO exhibit a more conservative convergence pattern. Among them, DDPG achieves the lowest steady-state relative optimality gap in the last 10\% of episodes, at 3.52\% $\pm$ 0.18\%; PPO follows with 4.25\% $\pm$ 0.64\%; and A2C and TRPO reach 6.05\% $\pm$ 0.31\% and 6.30\% $\pm$ 0.17\%, respectively. In terms of steady-state mean performance, DDPG reduces the optimality gap by 2.53, 2.78, and 0.73 percentage points relative to A2C, TRPO, and PPO, respectively, corresponding to relative reductions of 41.82\%, 44.13\%, and 17.18\%. PPO also reduces the gap by 1.80 and 2.05 percentage points relative to A2C and TRPO, respectively, corresponding to relative reductions of 29.75\% and 32.54\%.

In terms of sample efficiency, taking $\delta = 10\%$ as the threshold for the optimality gap, all four algorithms are able to reach the target, although with substantial differences in convergence speed. PPO is the first to reach the 10\% threshold at episode 328, followed closely by DDPG at episode 336, whereas A2C and TRPO reach the same threshold at episodes 506 and 695, respectively. Under the stricter 5\% threshold, only PPO (episode 398) and DDPG (episode 340) are able to meet the criterion, with DDPG entering the 5\% region earlier. These results indicate that PPO and DDPG possess stronger sample efficiency and steady-state approximation capability in this task. By contrast, although A2C and TRPO can stably enter the effective region, they fail to converge further to within 5\%, reflecting a more conservative search behavior in the DPMP representation space.

From the perspective of the trade-off between approximation depth and steady-state stability, PPO and DDPG exhibit different strengths. PPO achieves the lowest minimum moving-average (MA) optimality gap over the entire training process, at 1.21\%, which is the best among all four algorithms, indicating a stronger ability to approach the optimal policy during training. However, its standard deviation in the last 10\% of episodes is 0.64\%, which is higher than DDPG’s 0.18\%, implying more pronounced fluctuations in the later stage. By contrast, DDPG attains a minimum MA of 1.51\%, slightly higher than that of PPO, but achieves a lower steady-state mean and smaller fluctuations, indicating a performance profile with slightly weaker extreme-case approximation but stronger steady-state behavior. TRPO exhibits the smallest late-stage standard deviation (0.17\%), but its steady-state plateau remains relatively high (6.30\%), reflecting a stable yet conservative convergence pattern. A2C converges faster than TRPO (506 vs. 695), but its steady-state approximation depth remains clearly inferior to that of PPO and DDPG.

In summary, the answer to RQ2 from the vertical experiments is clear: under the DPMP representation, different reinforcement learning algorithms significantly affect convergence speed, approximation depth to the optimum, and steady-state fluctuations, but they do not alter the basic conclusion that DPMP can be learned stably and can enter an effective near-optimal region. More importantly, all four mainstream algorithms, namely A2C, TRPO, PPO, and DDPG, perform well under DPMP (with pass rates of 100\% in the last 10\% of episodes), indicating that the advantages of DPMP are not tied to any specific algorithm (such as PPO), but instead exhibit good cross-algorithm compatibility and engineering generality. Among them, PPO and DDPG achieve stronger optimality-approximation performance in this task, whereas A2C and TRPO provide more conservative but stable convergence behavior.

\begin{figure}[H]
    \centering
    \includegraphics[width=0.65\linewidth]{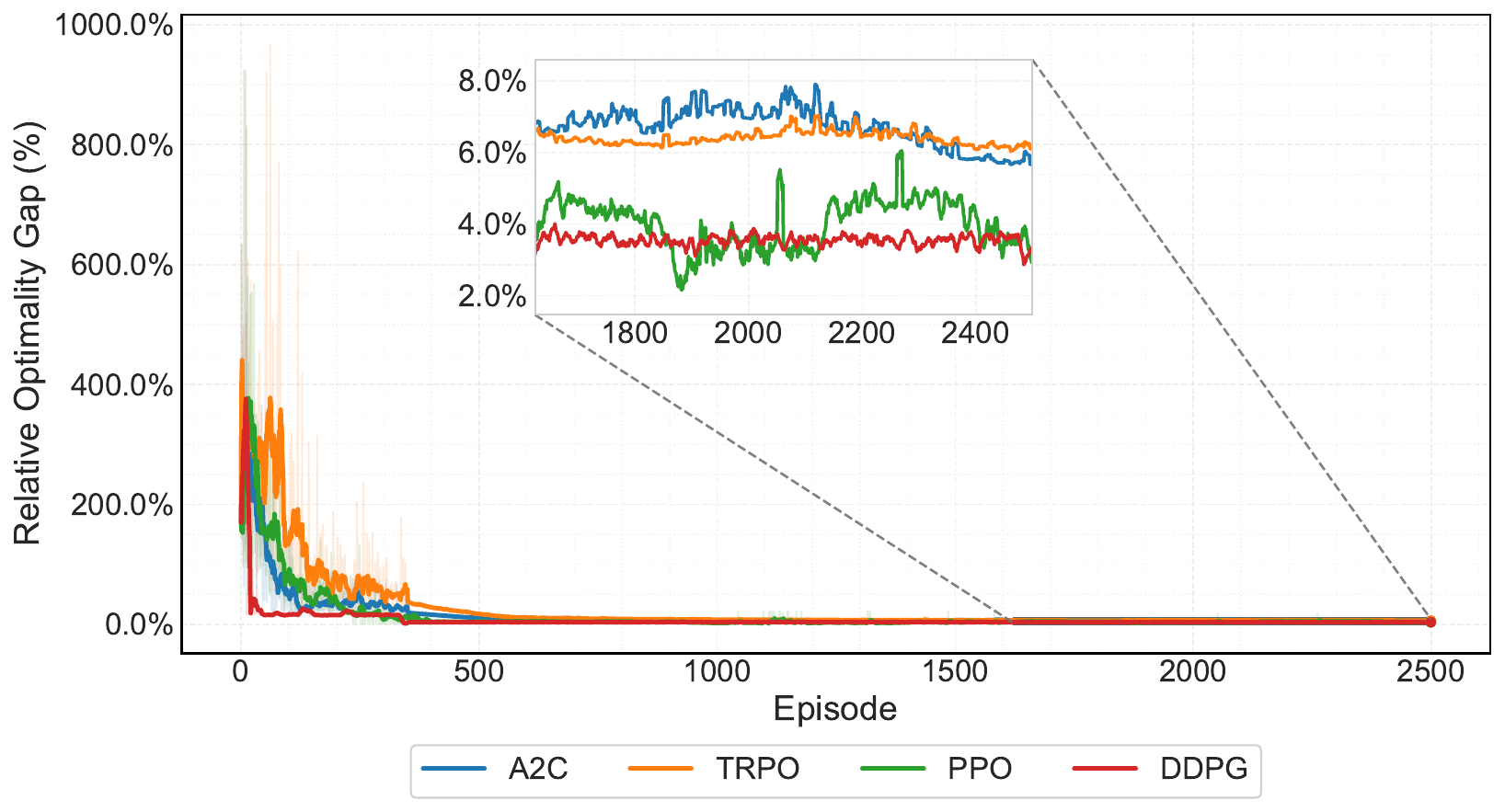}
    \caption{Optimality-gap curves of the four algorithms.}
    \label{fig:four_algorithms_relative_opt_gap}
\end{figure}

\begin{table}[H]
\centering
\caption{Statistics of the optimality-gap curves in the vertical experiments.}
\label{tab:vertical_opt_gap_stats}
\small
\renewcommand{\arraystretch}{1.15}
\setlength{\tabcolsep}{4pt}

\begin{tabularx}{\linewidth}{
C{1.3cm}
C{2.7cm}
C{2.0cm}
C{2.0cm}
C{2.3cm}
C{1.9cm}
}
\toprule
Method &
\makecell[c]{Final steady-state\\optimality gap\\(last 10\%,\\mean $\pm$ std)} &
\makecell[c]{Episode reaching\\the 10\%\\gap threshold} &
\makecell[c]{Episode reaching\\the 5\%\\gap threshold} &
\makecell[c]{Best gap\\(minimum MA\\over all episodes)} &
\makecell[c]{Pass rate in the\\last 10\%\\($\le$10\%)} \\
\midrule
A2C  & $6.05\%\pm0.31\%$ & 506 & --  & $5.66\%$ & $100.00\%$ \\
TRPO & $6.30\%\pm0.17\%$ & 695 & --  & $6.03\%$ & $100.00\%$ \\
PPO  & $4.25\%\pm0.64\%$ & 328 & 398 & $1.21\%$ & $100.00\%$ \\
DDPG & $3.52\%\pm0.18\%$ & 336 & 340 & $1.51\%$ & $100.00\%$ \\
\bottomrule
\end{tabularx}
\end{table}

\subsection{Multi-Agent Exploitability Assessment of DPMP}

\subsubsection{Research Question}

The previous subsection completed the first-level assessment of DPMP. The results showed that DPMP can significantly improve optimality approximation and exhibits good compatibility with mainstream algorithms such as A2C, TRPO, PPO, and DDPG, with PPO and DDPG demonstrating better performance on this task. Accordingly, this section turns to the second-level validity assessment defined in Section 5. In the IEEE 39-bus multi-agent market environment, we adopt the DPMP-PPO multi-agent algorithm to conduct electricity market simulation and exploitability assessment in a multi-agent setting. Specifically, this section aims to address the following research question.

\textbf{RQ3:} Under the general-sum market environment and strategic interaction setting of the IEEE 39-bus system, do the multi-agent simulation results obtained by training DPMP-PPO still admit significant room for unilateral improvement? In other words, is the exploitability sufficiently small for the resulting strategy profile to be regarded as a stable outcome in the sense of an approximate $\varepsilon$-Nash equilibrium?

\subsubsection{Experimental Setting}

This section constructs an IEEE 39-bus network-constrained market environment and adopts the DPMP-PPO strategy representation and learning framework. The environment explicitly incorporates network power flows, transmission-line constraints, unit start-up/shut-down constraints, and ramping constraints, so that the exploitability assessment is carried out in a multi-agent setting that more closely resembles actual market operation. Detailed environment settings are provided in Appendix C.

\subsubsection{Experiments and Results Analysis of the Multi-Agent Exploitability Assessment}

To answer RQ3, this section adopts exploitability as the evaluation metric and estimates the profit improvement brought by unilateral deviation via an RL-based approximate best response. When $\hat{\mathcal{E}}(\pi)$ is sufficiently small, the simulation result may be regarded as an approximate $\varepsilon$-Nash equilibrium.

\noindent
\textbf{(1) Convergence Pattern of Baseline Multi-Agent Training}

\begin{figure}[H]
    \centering
    \includegraphics[width=0.65\linewidth]{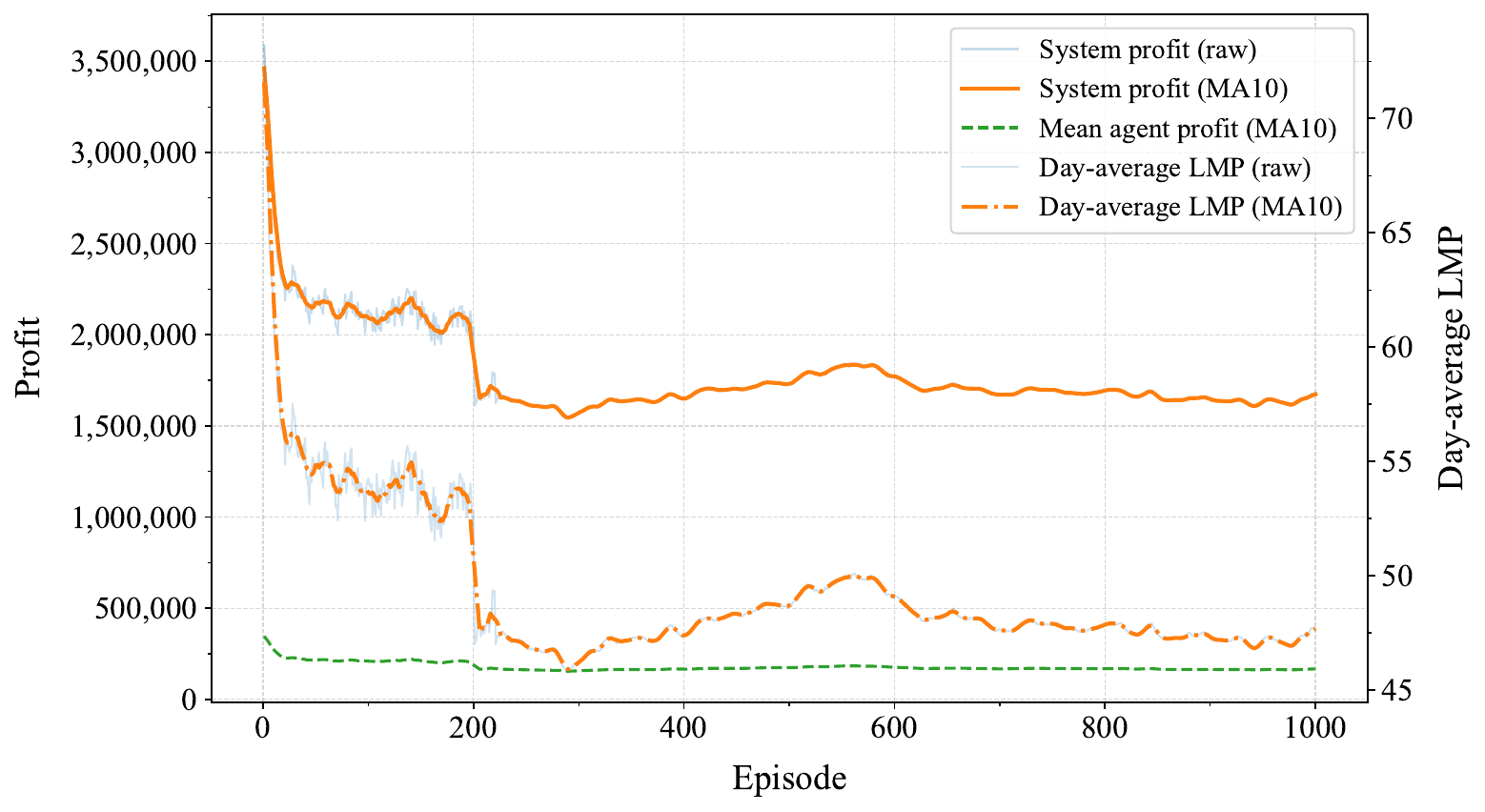}
    \caption{Multi-agent Baseline Training Convergence Overview.}
    \label{fig:baseline_training_overview}
\end{figure}

Figure~\ref{fig:baseline_training_overview} presents an overview of the convergence trajectories of system profit, average generator profit, and average daily LMP in the baseline DPMP-PPO multi-agent training. Overall, the training dynamics exhibit a clear two-stage pattern:

\begin{enumerate}[label=\roman*.]

    \item \textbf{Rapid adjustment stage (approximately 0--200 episodes):} Both system profit (MA10) and average daily LMP (MA10) show a pronounced downward trend, indicating that during early exploration the agents actively adjust their bid structures, rapidly shifting the market from a high-price, high-profit nonstationary region to a more competitive price regime. The raw curves fluctuate substantially in this stage, reflecting the strong nonstationarity induced by strategic interactions among multiple agents.
    
    \item \textbf{Steady-state stage (approximately 200--1000 episodes):} System profit (MA10) drifts slowly within a narrow band of approximately $1.6\sim1.7\times10^6$, while average daily LMP (MA10) stabilizes within a low-volatility range of about $46$--$47$ \$; average generator profit (MA10) is maintained correspondingly at roughly $1.6\sim1.7\times10^5$ \$. The joint stabilization of prices and profits indicates that, under the combined effects of practical constraints such as network constraints, unit start-up/shut-down, and ramping, strategic interactions have entered a relatively stable behavioral regime. This provides a necessary prerequisite for the subsequent exploitability assessment, namely, evaluating the gain from unilateral deviation in the vicinity of a near-stationary strategy profile rather than during the strongly nonstationary phase of training.

\end{enumerate}

\noindent
\textbf{(2) Agent exploitability}

\begin{figure}[H]
    \centering
    \includegraphics[width=0.65\linewidth]{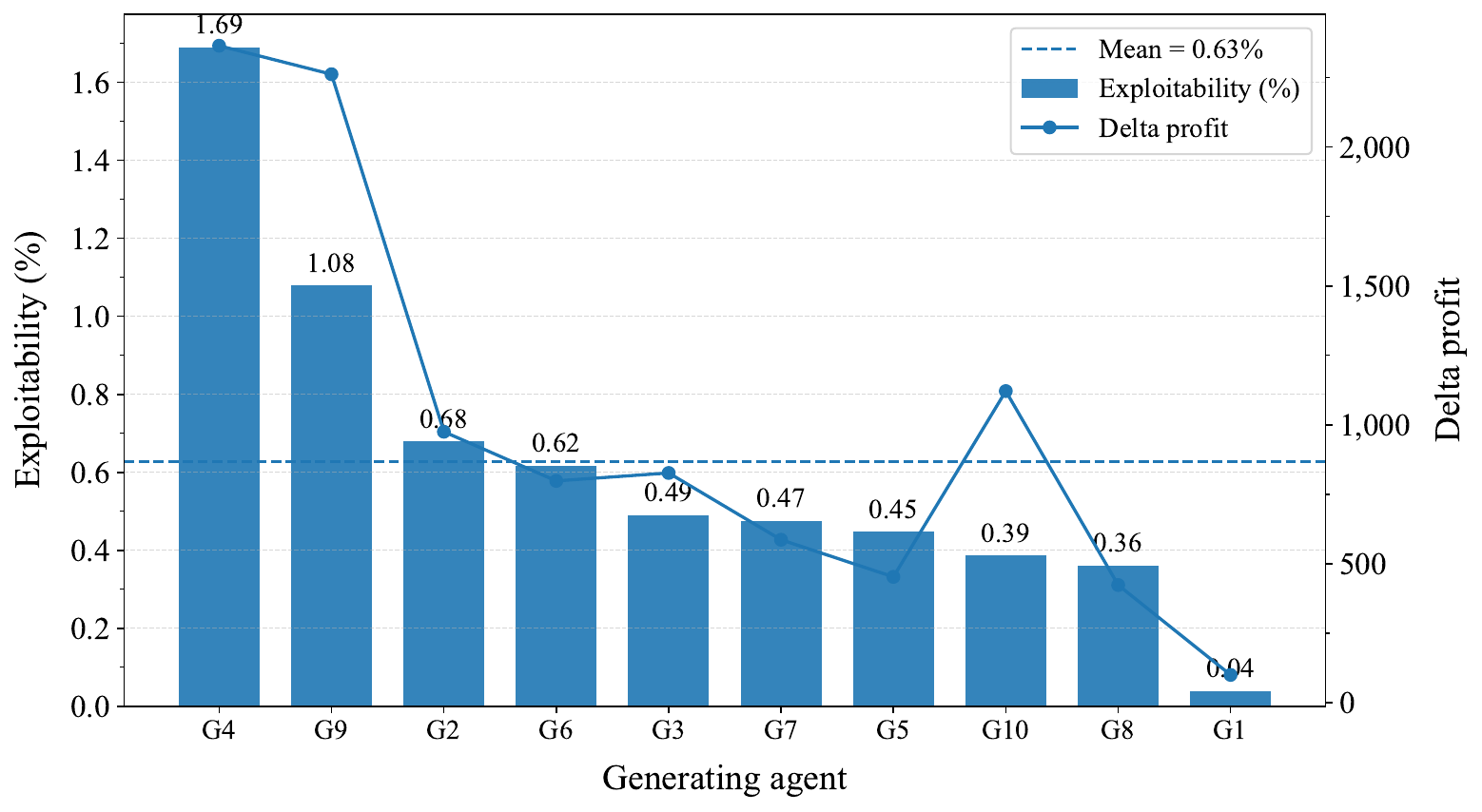}
    \caption{ Agent-wise Exploitability under DPMP-PPO Baseline Profile.}
    \label{fig:agent_exploitability_bar}
\end{figure}

Figure~\ref{fig:agent_exploitability_bar} and Table~\ref{tab:multiagent_exploitability_results} report, under the baseline DPMP-PPO simulation outcome, the unilateral deviation gain $\Delta_i$ and its normalized indicator $\mathrm{exploitability\ pct}$ obtained for each generator agent after applying the procedure of “freezing the other agents’ policies and retraining a single-agent RL approximate best response”.

The results exhibit a highly sparse pattern of profit improvement, together with a low overall mean level:

\begin{enumerate}[label=\roman*.]
    \item \textbf{Most agents are non-exploitable:}
    Among the 10 agents, 6 satisfy $\Delta_i=0$ and $\mathrm{exploitability\ pct}=0$
    (Agents 1/2/3/4/8/10). This means that, with the other agents’ policies frozen, these agents are unable to obtain a better strategy or achieve any additional profit improvement even after retraining an approximate best response, indicating strong unilateral stability.
    
    \item \textbf{Nonzero exploitability is concentrated in a few agents and remains limited in magnitude:}
    Only 4 agents exhibit nonzero $\mathrm{exploitability\ pct}$ values:
    Agent 6 has $0.016\%$ ($\Delta\approx 24.74$),
    Agent 5 has $0.323\%$ ($\Delta\approx 328.61$),
    Agent 7 has $0.440\%$ ($\Delta\approx 530.63$), and
    Agent 9 has $1.266\%$ ($\Delta\approx 2548.32$).
    The maximum value is contributed by Agent 9, namely
    $\hat{\mathcal{E}}(\pi)=1.266\%$, while all remaining agents are below $0.5\%$.
    
    \item \textbf{The overall mean remains low:}
    The average $\mathrm{exploitability\ pct}$ across the 10 agents is approximately $0.20\%$ (indicated by the dashed line in the figure).
\end{enumerate}

Taken together, both the maximum value $\hat{\mathcal{E}}(\pi)$ and the mean $\mathrm{exploitability\ pct}$ indicate that the overall room for unilateral improvement is small, and the simulation outcome does not exhibit a pronounced nonequilibrium pattern that can be substantially exploited. Therefore, the resulting multi-agent simulation outcome can, overall, be regarded as a stable state in the sense of an approximate $\varepsilon$-Nash equilibrium.

\begin{table*}[!t]
\centering
\caption{Results of the Multi-Agent Exploitability Assessment}
\label{tab:multiagent_exploitability_results}
\small
\renewcommand{\arraystretch}{1.12}
\setlength{\tabcolsep}{4pt}

\resizebox{\textwidth}{!}{
\begin{tabular}{c c c c c c c c}
\toprule
\makecell[c]{agent\\id} &
\makecell[c]{baseline\\profit} &
\makecell[c]{br\ profit} &
\makecell[c]{delta\\ profit} &
\makecell[c]{exploitabil\\ity\ rel} &
\makecell[c]{exploitabil\\ity\ pct} &
\makecell[c]{baseline\ \\total\ profit} &
\makecell[c]{br\ profile\\ total\ profit} \\
\midrule
1  & 261889.5264 & 261517.1334 & 0          & 0          & 0         & 1703768.7707 & 1700143.12 \\
2  & 144693.6381 & 144641.6330 & 0          & 0          & 0         & 1703768.7707 & 1703906.508 \\
3  & 167444.5666 & 167372.2242 & 0          & 0          & 0         & 1703768.7707 & 1699983.681 \\
4  & 143990.0787 & 140141.9598 & 0          & 0          & 0         & 1703768.7707 & 1706268.756 \\
5  & 101836.5862 & 102165.2002 & 328.6139737 & 0.003226875 & 0.322687539 & 1703768.7707 & 1708965.378 \\
6  & 153936.7463 & 153961.4858 & 24.73949964 & 0.000160712 & 0.016071211 & 1703768.7707 & 1705455.52 \\
7  & 120577.9228 & 121108.5547 & 530.6319477 & 0.004400738 & 0.440073884 & 1703768.7707 & 1686473.76 \\
8  & 116157.4190 & 116051.2556 & 0          & 0          & 0         & 1703768.7707 & 1700793.683 \\
9  & 201244.0092 & 203792.3274 & 2548.318175 & 0.012662828 & 1.266282751 & 1703768.7707 & 1723780.593 \\
10 & 291998.2138 & 291519.2364 & 0          & 0          & 0         & 1703768.7707 & 1703631.063 \\
\bottomrule
\end{tabular}
}
\end{table*}

\noindent
\textbf{(3) System-level implications}

Table~\ref{tab:multiagent_exploitability_results} also reports the baseline system-wide total profit, $\mathrm{baseline\ total\ profit}$ (approximately $1.7038\times10^6$), together with $\mathrm{br\ profile\ total\ profit}$ after introducing a single-agent best-response (BR) deviation. A comparison of the two shows that, even when the deviation is made by the agent with the highest exploitability, the resulting change in system-level profit remains confined to a narrow range:

\begin{enumerate}[label=\roman*.]
    \item \textbf{The largest positive change} occurs under Agent 9's BR deviation, where the system-wide total profit increases by approximately $+1.17\%$
    ($1.7038 \rightarrow 1.7238$ million \$).
    
    \item \textbf{The largest negative change} occurs under Agent 7's BR deviation, where the system-wide total profit decreases by approximately $-1.02\%$
    ($1.7038 \rightarrow 1.6865$ million \$).
\end{enumerate}

This indicates that, although a very small unilateral deviation gain still exists—that is, the outcome is not a strict Nash equilibrium—the resulting profit improvement is manifested mainly as a marginal individual-level gain, rather than inducing any systemic redistribution or sharp change in the pricing mechanism. This point is particularly important for electricity market mechanism analysis and evaluation: it implies that subsequent comparisons across different market mechanisms primarily reflect steady-state differences caused by the mechanisms themselves (e.g., constraints, market clearing, and pricing rules), rather than uncontrolled bias introduced by the fact that the outcomes are not strict Nash equilibria.

\section{Conclusions}

This study focuses on two key issues in reinforcement learning agent-based simulation (RL-ABS) for electricity market research. First, existing post-processing mappings for constraint handling, such as sorting, clipping, and projection, disrupt the correspondence between policy outputs and executed bids, thereby leading to gradient distortion and objective mismatch during training. Second, existing studies typically lack quantitative tests of the equilibrium validity of simulation results, making it difficult for simulation-based conclusions to directly support electricity market mechanism analysis. To address these two issues, this paper proposes the continuously differentiable, injective, and invertible Dual-Positive Monotone Parameterization (DPMP), together with a two-level validity assessment framework composed of the optimality gap and exploitability.

The experimental results show that, in the single-agent setting, DPMP can approximate the theoretical optimal profit more stably than post-processing mapping baselines based on sorting, projection, and clipping, while significantly reducing the steady-state optimality gap. In particular, the late-stage relative optimality gap can be reduced from about 30\% to about 3\%. Furthermore, this advantage remains consistent across multiple mainstream deep reinforcement learning algorithms, indicating that the improvement arises primarily from the optimization of action representation rather than from reliance on specific algorithmic heuristics. In the IEEE 39-bus network-constrained multi-agent day-ahead market simulation, this study further confirms that the simulation results exhibit low exploitability, suggesting that they can be regarded as stable solutions in the sense of an approximate $\varepsilon$-Nash equilibrium.

Overall, the main contribution of this work lies not merely in proposing a new bid parameterization and mapping method, but more importantly in enhancing the interpretability and credibility of the results when RL-ABS is used as a tool for electricity market research. On the one hand, DPMP reduces the learning–execution bias caused by post-processing mappings at the level of action representation. On the other hand, the two-level validity assessment framework elevates simulation outcomes from mere convergence of training curves to market outcomes with quantifiable equilibrium significance. This provides a more reliable methodological foundation for using RL-ABS to conduct electricity market rule comparison, market power analysis, and mechanism evaluation. In addition, although the proposed method is developed and validated in the context of electricity market RL-ABS, its underlying ideas may also be informative for other continuous decision problems with similar constraint structures, especially in adjacent energy-system applications involving monotonicity, boundedness, or piecewise structural constraints. Future work may further investigate the applicability and generalizability of DPMP and the proposed validity assessment framework in such settings.

% \section*{Acknowledgement}
% This work was supported by

%% If you have bibdatabase file and want bibtex to generate the bibitems, please use

\bibliographystyle{elsarticle-num}
\bibliography{References.bib}

@article{glismann2021ancillary,
  title={Ancillary Services Acquisition Model: Considering market interactions in policy design},
  author={Glismann, Samuel},
  journal={Applied Energy},
  volume={304},
  pages={117697},
  year={2021},
  publisher={Elsevier}
}

@inproceedings{pan2023multi,
  title={A multi-agent simulation model considering the bounded rationality of market participants: an example of GENCOs participation in the electricity spot market},
  author={Pan, Zhanhua and Jing, Zhaoxia and Ji, Tianyao and Song, Yuhui},
  booktitle={International Workshop on Multi-Agent Systems and Agent-Based Simulation},
  pages={129--145},
  year={2023},
  organization={Springer}
}

@article{sridhar2024residential,
  title={Residential consumer enrollment in demand response: An agent based approach},
  author={Sridhar, Araavind and Honkapuro, Samuli and Ruiz, Fredy and Stoklasa, Jan and Annala, Salla and Wolff, Annika},
  journal={Applied Energy},
  volume={374},
  pages={123988},
  year={2024},
  publisher={Elsevier}
}

@article{ventosa2005electricity,
  title={Electricity market modeling trends},
  author={Ventosa, Mariano and Ba{\i}llo, Alvaro and Ramos, Andr{\'e}s and Rivier, Michel},
  journal={Energy policy},
  volume={33},
  number={7},
  pages={897--913},
  year={2005},
  publisher={Elsevier}
}

@article{baillo2004optimal,
  title={Optimal offering strategies for generation companies operating in electricity spot markets},
  author={Baillo, Alvaro and Ventosa, Mariano and Rivier, Michel and Ramos, Andres},
  journal={IEEE Transactions on Power Systems},
  volume={19},
  number={2},
  pages={745--753},
  year={2004},
  publisher={IEEE}
}

@article{hobbs2000strategic,
  title={Strategic gaming analysis for electric power systems: An MPEC approach},
  author={Hobbs, Benjamin F and Metzler, Carolyn B and Pang, J-S},
  journal={IEEE transactions on power systems},
  volume={15},
  number={2},
  pages={638--645},
  year={2000},
  publisher={IEEE}
}

@article{ringler2016agent,
  title={Agent-based modelling and simulation of smart electricity grids and markets--a literature review},
  author={Ringler, Philipp and Keles, Dogan and Fichtner, Wolf},
  journal={Renewable and Sustainable Energy Reviews},
  volume={57},
  pages={205--215},
  year={2016},
  publisher={Elsevier}
}

@article{shafie2014stochastic,
  title={A stochastic multi-layer agent-based model to study electricity market participants behavior},
  author={Shafie-Khah, Miadreza and Catal{\~a}o, Jo{\~a}o PS},
  journal={IEEE Transactions on Power Systems},
  volume={30},
  number={2},
  pages={867--881},
  year={2014},
  publisher={IEEE}
}

@article{fraunholz2021advanced,
  title={Advanced price forecasting in agent-based electricity market simulation},
  author={Fraunholz, Christoph and Kraft, Emil and Keles, Dogan and Fichtner, Wolf},
  journal={Applied Energy},
  volume={290},
  pages={116688},
  year={2021},
  publisher={Elsevier}
}

@article{nanduri2007reinforcement,
  title={A reinforcement learning model to assess market power under auction-based energy pricing},
  author={Nanduri, Vishnuteja and Das, Tapas K},
  journal={IEEE transactions on Power Systems},
  volume={22},
  number={1},
  pages={85--95},
  year={2007},
  publisher={IEEE}
}

@article{song2025optimal,
  title={Optimal Bidding Framework for Integrated Renewable-Storage Plant in High-Dimensional Real-Time Markets},
  author={Song, Yuhao and Huang, Shaowei and Chen, Laijun and Cui, Sen and Mei, Shengwei},
  journal={Sustainability},
  volume={17},
  number={18},
  pages={8159},
  year={2025},
  publisher={MDPI}
}

@article{yu2023reinforcement,
  title={A reinforcement-probability Bayesian approach for strategic bidding and market clearing for renewable energy sources with uncertainty},
  author={Yu, Liying and Wang, Peng and Zhang, Yang and Li, Ning and Cherkaoui, Rachid},
  journal={Journal of Cleaner Production},
  volume={429},
  pages={139403},
  year={2023},
  publisher={Elsevier}
}

@article{chandrakala2023multi,
  title={Multi-agent based modeling and learning approach for intelligent day-ahead bidding strategy in wholesale electricity market},
  author={Chandrakala, KRM Vijaya and Kiran, P},
  journal={Expert Systems with Applications},
  volume={233},
  pages={121014},
  year={2023},
  publisher={Elsevier}
}

@article{rahimiyan2010adaptive,
  title={An adaptive $ Q $-learning algorithm developed for agent-based computational modeling of electricity market},
  author={Rahimiyan, Morteza and Mashhadi, Habib Rajabi},
  journal={IEEE Transactions on Systems, Man, and Cybernetics, Part C (Applications and Reviews)},
  volume={40},
  number={5},
  pages={547--556},
  year={2010},
  publisher={IEEE}
}

@article{weng2025optimizing,
  title={Optimizing bidding strategy in electricity market based on graph convolutional neural network and deep reinforcement learning},
  author={Weng, Haoen and Hu, Yongli and Liang, Min and Xi, Jiayang and Yin, Baocai},
  journal={Applied Energy},
  volume={380},
  pages={124978},
  year={2025},
  publisher={Elsevier}
}

@article{yin2024multi,
  title={Multi-agent deep reinforcement learning for simulating centralized double-sided auction electricity market},
  author={Yin, Baocai and Weng, Haoen and Hu, Yongli and Xi, Jiayang and Ding, Pinggang and Liu, Jia},
  journal={IEEE Transactions on Power Systems},
  volume={40},
  number={1},
  pages={518--529},
  year={2024},
  publisher={IEEE}
}

@article{rokhforoz2023multi,
  title={Multi-agent reinforcement learning with graph convolutional neural networks for optimal bidding strategies of generation units in electricity markets},
  author={Rokhforoz, Pegah and Montazeri, Mina and Fink, Olga},
  journal={Expert Systems with Applications},
  volume={225},
  pages={120010},
  year={2023},
  publisher={Elsevier}
}

@article{liang2020agent,
  title={Agent-based modeling in electricity market using deep deterministic policy gradient algorithm},
  author={Liang, Yanchang and Guo, Chunlin and Ding, Zhaohao and Hua, Huichun},
  journal={IEEE transactions on power systems},
  volume={35},
  number={6},
  pages={4180--4192},
  year={2020},
  publisher={IEEE}
}

@article{zhang2024game,
  title={Game bidding and benefit allocation strategy for virtual power plants with multiple new market entities based on multi-agent reinforcement learning},
  author={ZHANG, JH and Zhang, Y and Wang, X and JIANG, CW and WANG, LL},
  journal={Power System Technology},
  pages={1--12},
  year={2024}
}

@article{pan2025decision,
  title={Decision-making and cost models of generation company agents for supporting future electricity market mechanism design based on agent-based simulation},
  author={Pan, Zhanhua and Jing, Zhaoxia},
  journal={Applied Energy},
  volume={391},
  pages={125881},
  year={2025},
  publisher={Elsevier}
}

@article{jiang2024optimal,
  title={Optimal bidding strategy for the price-maker virtual power plant in the day-ahead market based on multi-agent twin delayed deep deterministic policy gradient algorithm},
  author={Jiang, Yuzheng and Dong, Jun and Huang, Hexiang},
  journal={Energy},
  volume={306},
  pages={132388},
  year={2024},
  publisher={Elsevier}
}

@book{sutton1998reinforcement,
  title={Reinforcement learning: An introduction},
  author={Sutton, Richard S and Barto, Andrew G and others},
  volume={1},
  number={1},
  year={1998},
  publisher={MIT press Cambridge}
}

@book{zhao2025mathematical,
  title={Mathematical foundations of reinforcement learning},
  author={Zhao, Shiyu},
  year={2025},
  publisher={Springer Nature}
}

@article{wu2024intelligent,
  title={Intelligent strategic bidding in competitive electricity markets using multi-agent simulation and deep reinforcement learning},
  author={Wu, Jiahui and Wang, Jidong and Kong, Xiangyu},
  journal={Applied Soft Computing},
  volume={152},
  pages={111235},
  year={2024},
  publisher={Elsevier}
}

@article{lohndorf2013optimizing,
  title={Optimizing trading decisions for hydro storage systems using approximate dual dynamic programming},
  author={L{\"o}hndorf, Nils and Wozabal, David and Minner, Stefan},
  journal={Operations Research},
  volume={61},
  number={4},
  pages={810--823},
  year={2013},
  publisher={INFORMS}
}

@misc{manual202111,
  title={11: Energy and Ancillary Services Market Operations Revision: 122},
  author={Manual, PJMPJM},
  year={2021},
  publisher={PJM, Norristown, PA, USA}
}

@inproceedings{fujita2018clipped,
  title={Clipped action policy gradient},
  author={Fujita, Yasuhiro and Maeda, Shin-ichi},
  booktitle={International conference on machine learning},
  pages={1597--1606},
  year={2018},
  organization={PMLR}
}

@article{yu2023finding,
  title={Finding Nash equilibrium based on reinforcement learning for bidding strategy and distributed algorithm for ISO in imperfect electricity market},
  author={Yu, Liying and Wang, Peng and Chen, Zhe and Li, Dewen and Li, Ning and Cherkaoui, Rachid},
  journal={Applied Energy},
  volume={350},
  pages={121704},
  year={2023},
  publisher={Elsevier}
}

@article{lanctot2019openspiel,
  title={OpenSpiel: A framework for reinforcement learning in games},
  author={Lanctot, Marc and Lockhart, Edward and Lespiau, Jean-Baptiste and Zambaldi, Vinicius and Upadhyay, Satyaki and P{\'e}rolat, Julien and Srinivasan, Sriram and Timbers, Finbarr and Tuyls, Karl and Omidshafiei, Shayegan and others},
  journal={arXiv preprint arXiv:1908.09453},
  year={2019}
}

@article{de2010isotone,
  title={Isotone optimization in R: pool-adjacent-violators algorithm (PAVA) and active set methods},
  author={De Leeuw, Jan and Hornik, Kurt and Mair, Patrick},
  journal={Journal of statistical software},
  volume={32},
  pages={1--24},
  year={2010}
}
%\end{thebibliography}

\noindent
\textbf{Appendix A.}

\noindent
\textbf{A.1 Proof of Necessary Condition 1 (NC1) for Post-Processing Operations}

\textbf{Necessary Condition 1 (NC1):} The post-processing mapping $h$ should satisfy
\[
\forall a_0,\ \mathbb{P}\bigl(h(x)=a_0 \mid s\bigr)=0
\]

\textbf{Proof by contradiction:} Assume that under the post-processing mapping $h$, there exists some $a_0$ such that

\begin{equation}
\mathbb{P}\bigl(h(x)=a_0 \mid s\bigr)>0
\label{eq:appendix_nc1_contra}
\end{equation}

Then the distribution of the executed action $a$ is no longer an ordinary probability density function; that is, $\pi_\theta(a \mid s)$ becomes a generalized function. Under the standard continuous-action policy gradient implementations considered in this paper, $\log \pi_\theta(a \mid s)$ and its gradient, both of which are defined on the basis of an ordinary continuous density, would no longer be objects that can be computed directly and stably. This contradicts the standard continuous-action policy gradient setting considered in this paper. Therefore, the assumption is false.

\noindent
\textbf{A.2 Proof of Necessary Condition 2 (NC2) for Post-Processing Operations}

\textbf{Necessary Condition 2 (NC2):} On the non-redundant action space $Z$, the non-redundant mapping $g$ should be injective, i.e.,

\[
\mathbb{P}\bigl(\exists z' \neq z : g(z') = g(z) \mid s\bigr)=0.
\]

\textbf{Proof:}

Let $z \in \mathbb{R}^d$ have probability density function $\tilde{\mu}_\theta(z \mid s)$. Suppose that, in a neighborhood of some $a$, the equation $g(z)=a$ has $m$ distinct solutions $\{z_k(a)\}_{k=1}^m$, and that the Jacobian is invertible at each solution (i.e., $\det J_g(z_k) \neq 0$).

By the inverse function theorem, if $\det J_g(z_k) \neq 0$, then there exists a local inverse mapping $z=\varphi_k(a)$ in a neighborhood of $z_k$, and $g$ is one-to-one between this local neighborhood and its image. Decomposing the $z$-space into these pairwise disjoint locally invertible neighborhoods and applying the multivariate change-of-variables formula to each branch yields

\begin{equation}
\int_{g(U_k)} \pi_\theta(a \mid s)\, da
=
\int_{U_k} \tilde{\mu}_\theta(z \mid s)\, dz
=
\int_{g(U_k)} \tilde{\mu}_\theta\bigl(\varphi_k(a) \mid s\bigr)\left|\det J_{\varphi_k}(a)\right|\, da
\label{eq:appendix_nc2_change_of_variables}
\end{equation}

Since $J_{\varphi_k}(a)=J_g\bigl(z_k(a)\bigr)^{-1}$, we have $\left|\det J_{\varphi_k}\right|=1/\left|\det J_g\right|$. Summing over all branches gives the executed probability density function:

\begin{equation}
\pi_\theta(a \mid s)
=
\sum_{k=1}^m
\frac{\tilde{\mu}_\theta\bigl(z_k(a) \mid s\bigr)}
{\left|\det J_g\bigl(z_k(a)\bigr)\right|}
\label{eq:appendix_nc2_density_sum}
\end{equation}

Define
\begin{equation}
w_k(a):=
\frac{\tilde{\mu}_\theta\bigl(z_k(a) \mid s\bigr)}
{\left|\det J_g\bigl(z_k(a)\bigr)\right|},
\qquad
\pi_\theta(a \mid s)=\sum_{k=1}^m w_k(a)
\label{eq:appendix_nc2_wk_def}
\end{equation}

Then
\begin{equation}
\nabla_\theta \log \pi_\theta(a \mid s)
=
\frac{1}{\sum_{j=1}^m w_j(a)}
\sum_{k=1}^m \nabla_\theta w_k(a)
=
\sum_{k=1}^m \alpha_k(a)\,\nabla_\theta \log w_k(a)
\label{eq:appendix_nc2_logpi_grad}
\end{equation}

where the mixing weights are
\begin{equation}
\alpha_k(a):=
\frac{w_k(a)}{\sum_{j=1}^m w_j(a)},
\qquad
\sum_{k=1}^m \alpha_k(a)=1
\label{eq:appendix_nc2_alpha_def}
\end{equation}

Moreover,
\begin{equation}
\nabla_\theta \log w_k(a)
=
\nabla_\theta \log \tilde{\mu}_\theta\bigl(z_k(a)\mid s\bigr)
\label{eq:appendix_nc2_logwk_grad}
\end{equation}

Equation \eqref{eq:appendix_nc2_logpi_grad} shows that the correct quantity $\nabla_\theta \log \pi_\theta(a \mid s)$ is a weighted average of $\nabla_\theta \log w_k(a)$ over all branches. However, in the implementation of stochastic policy gradient methods, only one sampled $z$ is observed, which is equivalent to obtaining only one term, namely $\nabla_\theta \log \tilde{\mu}_\theta(z \mid s)$. The contributions from the other branches are not generated automatically, nor can the weights $\alpha_k(a)$ be computed. Therefore, unless $m=1$, i.e., unless the mapping is injective, the gradient obtained by backpropagation along a single path is not the same object as that in \eqref{eq:appendix_nc2_logpi_grad}, which in turn leads to an objective mismatch in stochastic policy gradient optimization.

\noindent
\textbf{A.3 Proof of Necessary Condition 3 (NC3) for Post-Processing Operations}

\textbf{Necessary Condition 3 (NC3):} The non-redundant post-processing mapping $g$ must be locally invertible at almost every sampled $z$, i.e.,

\[
\mathbb{P}\bigl(\det J_g(z)\neq 0 \mid s\bigr)=1
\]

\textbf{Proof (by contradiction):}

Assume, to the contrary, that there exist points sampled with positive probability such that

\begin{equation}
\det J_g(z)=0
\label{eq:appendix_nc3_singular}
\end{equation}

Then one may choose a sampled singular point $z_0$ such that

\begin{equation}
\operatorname{rank} J_g(z_0)=r<d
\label{eq:appendix_nc3_rank}
\end{equation}

By linear algebra, there exists a nonzero vector $v\neq 0$ such that

\begin{equation}
J_g(z_0)v=0
\label{eq:appendix_nc3_nullvec}
\end{equation}

Since $g$ is differentiable at $z_0$, for any sufficiently small $\varepsilon$, the first-order Taylor expansion gives

\begin{equation}
g(z_0+\varepsilon v)
=
g(z_0)+J_g(z_0)(\varepsilon v)+o(\varepsilon)
\label{eq:appendix_nc3_taylor}
\end{equation}

Combining this with \eqref{eq:appendix_nc3_nullvec} yields

\begin{equation}
g(z_0+\varepsilon v)=g(z_0)+o(\varepsilon)
\label{eq:appendix_nc3_collapse}
\end{equation}

Equation \eqref{eq:appendix_nc3_collapse} indicates that the perturbation of the raw action along direction $v$ is of order $O(\varepsilon)$, whereas the first-order variation in the executed action is zero, leaving only a higher-order infinitesimal term $o(\varepsilon)$. In other words, there exists a nonzero perturbation direction in a neighborhood of $z_0$ whose variation cannot be transmitted to the executed action at the first-order level, implying a collapse of the local sensitivity of the mapping $g$.

However, the standard continuous-action policy gradient implementations considered in this paper require that, at almost every sampled action point, small perturbations in the raw action can be transmitted stably to the executed action, so that the executed-action distribution and its associated gradient objects can be characterized consistently and stably. If such singular points exist in regions visited with positive probability, then different perturbations of the raw action become locally indistinguishable at the executed-action level, leading to degenerate gradient propagation and violating the basic requirements imposed by our framework on the post-processing mapping. Therefore, singular points cannot occur in regions sampled with positive probability, and NC3 is thus a necessary condition.

\noindent
\textbf{Appendix B Feasibility Analysis of the DPMP Mapping}

This appendix proves that, after removing the scale redundancy in the generation-output component, DPMP satisfies NC1--NC3 proposed in Section 3.

Let

\begin{equation}
\Delta p := p_{\max}-p_{\min}>0
\label{eq:appendixB_delta_p}
\end{equation}

\noindent
\textbf{B.1 Non-redundant action space and mapping under DPMP}

As shown in Section 3, since $\pi(cr)=\pi(r)$ holds for any $c>0$, the variable $r$ contains a one-dimensional scale redundancy. Let

\begin{equation}
\bar{\lambda}=(\lambda_1,\ldots,\lambda_{K-1})^\top,
\qquad
\lambda_K=1-\sum_{i=1}^{K-1}\lambda_i
\label{eq:appendixB_lambda_def}
\end{equation}

and define the non-redundant action space as

\begin{equation}
\mathcal{Z}^{\circ}
=
\left\{
(\bar{\lambda},w)
\;\middle|\;
\lambda_i>0,\ \sum_{i=1}^{K-1}\lambda_i<1,\ w_i>0
\right\}
\label{eq:appendixB_z_interior}
\end{equation}

Define the non-redundant output space as

\begin{equation}
\tilde{\mathcal{B}}_{K}^{\circ}
=
\left\{
(\tilde{Q},p)\in\mathbb{R}^{K-1}\times\mathbb{R}^{K}
\;\middle|\;
0<Q_1<\cdots<Q_{K-1}<Q_{\max},\ 
p_{\min}<p_1<\cdots<p_K<p_{\max}
\right\}
\label{eq:appendixB_Btilde_interior}
\end{equation}

where

\begin{equation}
\tilde{Q}:=(Q_1,\ldots,Q_{K-1})^\top
\label{eq:appendixB_Qtilde_def}
\end{equation}

Accordingly, the non-redundant DPMP mapping is given by

\begin{equation}
\widetilde{\Phi}:\mathcal{Z}^{\circ}\to\tilde{\mathcal{B}}_{K}^{\circ},
\qquad
\widetilde{\Phi}(\bar{\lambda},w)=(\tilde{Q},p)
\label{eq:appendixB_Phi_def}
\end{equation}

\noindent
\textbf{B.2 $\widetilde{\Phi}$ satisfies NC2: $\widetilde{\Phi}$ is injective on $\mathcal{Z}^{\circ}$.}

\textbf{Proof:}

Suppose that

\begin{equation}
\widetilde{\Phi}(\bar{\lambda},w)=\widetilde{\Phi}(\bar{\lambda}',w')
\label{eq:appendixB_inj_assume}
\end{equation}

Then

\begin{equation}
\tilde{Q}=\tilde{Q}',
\qquad
p=p'
\label{eq:appendixB_inj_equal_outputs}
\end{equation}

First consider the generation-output component. Since

\begin{equation}
Q_1=Q_{\max}\lambda_1,
\qquad
Q_i-Q_{i-1}=Q_{\max}\lambda_i,
\quad i=2,\ldots,K-1
\label{eq:appendixB_lambda_from_Q_forward}
\end{equation}

we can recover $\bar{\lambda}$ from $\tilde{Q}$ via
\begin{equation}
\lambda_1=\frac{Q_1}{Q_{\max}},
\qquad
\lambda_i=\frac{Q_i-Q_{i-1}}{Q_{\max}},
\quad i=2,\ldots,K-1
\label{eq:appendixB_lambda_from_Q_inverse}
\end{equation}

and
\begin{equation}
\lambda_K=1-\sum_{i=1}^{K-1}\lambda_i
\label{eq:appendixB_lambdaK_recover}
\end{equation}

Hence, $\tilde{Q}=\tilde{Q}' \Rightarrow \bar{\lambda}=\bar{\lambda}'$. Now consider the price component. Let

\begin{equation}
s_i=\sum_{j=1}^{i} w_j
\label{eq:appendixB_si_def}
\end{equation}

From
\begin{equation}
p_i=p_{\min}+\Delta p\bigl(1-e^{-s_i}\bigr)
\label{eq:appendixB_pi_from_si}
\end{equation}

we obtain the inverse relation
\begin{equation}
s_i=-\ln\!\left(1-\frac{p_i-p_{\min}}{\Delta p}\right)
\label{eq:appendixB_si_inverse}
\end{equation}

Therefore, $p=p' \Rightarrow s=s'$. Moreover, from
\begin{equation}
w_1=s_1,
\qquad
w_i=s_i-s_{i-1}\quad (i\ge 2)
\label{eq:appendixB_w_from_s}
\end{equation}

we obtain $w=w'$. Hence, $(\bar{\lambda},w)=(\bar{\lambda}',w')$, and injectivity is established.

\noindent
\textbf{B.3 $\widetilde{\Phi}$ satisfies NC3: $\widetilde{\Phi}$ is continuously differentiable throughout $\mathcal{Z}^{\circ}$, and its Jacobian matrix is nonsingular everywhere; therefore, it is locally invertible.}

\textbf{Proof:}

Let $L_{K-1}$ denote the upper-left $(K-1)\times (K-1)$ submatrix of $L$. Then
\begin{equation}
\tilde{Q}=Q_{\max}L_{K-1}\bar{\lambda}
\label{eq:appendixB_Q_block}
\end{equation}

Hence,
\begin{equation}
\frac{\partial \tilde{Q}}{\partial \bar{\lambda}}
=
Q_{\max}L_{K-1}
\label{eq:appendixB_dQ_dlambda}
\end{equation}

On the other hand,
\begin{equation}
p=p_{\min}\mathbf{1}+\Delta p\bigl(1-e^{-Lw}\bigr)
\label{eq:appendixB_p_block}
\end{equation}

Thus,
\begin{equation}
\frac{\partial p}{\partial w}
=
\Delta p\,\operatorname{diag}\!\bigl(e^{-Lw}\bigr)L
\label{eq:appendixB_dp_dw}
\end{equation}

Therefore, the Jacobian matrix of $\widetilde{\Phi}$ is
\begin{equation}
J_{\widetilde{\Phi}}
=
\begin{bmatrix}
Q_{\max}L_{K-1} & 0 \\
0 & \Delta p\,\operatorname{diag}\!\bigl(e^{-Lw}\bigr)L
\end{bmatrix}
\label{eq:appendixB_Jacobian}
\end{equation}

Here, both $L_{K-1}$ and $L$ are lower triangular with all diagonal entries equal to $1$, while all diagonal entries of $\operatorname{diag}(e^{-Lw})$ are strictly positive. Therefore,

\begin{equation}
\det\!\bigl(J_{\widetilde{\Phi}}\bigr)
=
(Q_{\max})^{K-1}(\Delta p)^K
\prod_{i=1}^{K} e^{-(Lw)_i}
>0
\label{eq:appendixB_det_positive}
\end{equation}

Hence, $J_{\widetilde{\Phi}}$ is nonsingular everywhere. By the inverse function theorem, $\widetilde{\Phi}$ is locally invertible throughout $\mathcal{Z}^{\circ}$, and NC3 holds.

\noindent
\textbf{B.4 $\widetilde{\Phi}$ satisfies NC1: if the policy admits a continuous density $\mu_\theta(z \mid s)$ on $\mathcal{Z}^{\circ}$, then after the non-redundant mapping $y=\widetilde{\Phi}(z)$, the output distribution still admits an ordinary density on $\widetilde{\Phi}(\mathcal{Z}^{\circ})$; therefore, no singular probability mass is generated on the boundary or on equality-constraint manifolds.}

\textbf{Proof:}

By B.1 and B.2, $\widetilde{\Phi}$ is a $C^1$ injective mapping on $\mathcal{Z}^{\circ}$, and its Jacobian determinant is nowhere zero. Therefore, the change-of-variables formula applies. The output density is thus given by

\begin{equation}
\rho_\theta(y \mid s)
=
\frac{\mu_\theta\bigl(\widetilde{\Phi}^{-1}(y)\mid s\bigr)}
{\left|\det J_{\widetilde{\Phi}}\bigl(\widetilde{\Phi}^{-1}(y)\bigr)\right|},
\qquad
y\in \widetilde{\Phi}(\mathcal{Z}^{\circ})
\label{eq:appendixB_rho_density}
\end{equation}

Moreover, the DPMP output always satisfies
\begin{equation}
0<Q_1<\cdots<Q_{K-1}<Q_{\max},
\qquad
p_{\min}<p_1<\cdots<p_K<p_{\max}
\label{eq:appendixB_output_strict}
\end{equation}

Therefore, the output can never fall on lower-dimensional boundary sets such as
\begin{equation}
Q_i=Q_{i-1},
\qquad
p_i=p_{i+1},
\qquad
p_i=p_{\min},
\qquad
p_i=p_{\max}
\label{eq:appendixB_boundary_sets}
\end{equation}

nor does DPMP compress an input region of positive volume into singular probability mass on the boundary, as sorting, clipping, or projection may do. Hence, NC1 holds. 

\noindent
\textbf{Appendix C Multi-Agent Simulation Environment Settings}

\noindent
\textbf{C.1 Physical Power System}

The simulation is conducted on the IEEE 39-bus system, as shown in Figure~\ref{fig:IEEE39}, which comprises 39 buses, 46 transmission lines, 10 generation units, and 18 load buses. The generating units are modeled according to Eq.~\eqref{eq:agent_mc}, and the corresponding parameters are reported in Table~\ref{tab:generator_parameters}:

\begin{figure}[H]
    \centering
    \includegraphics[width=\linewidth]{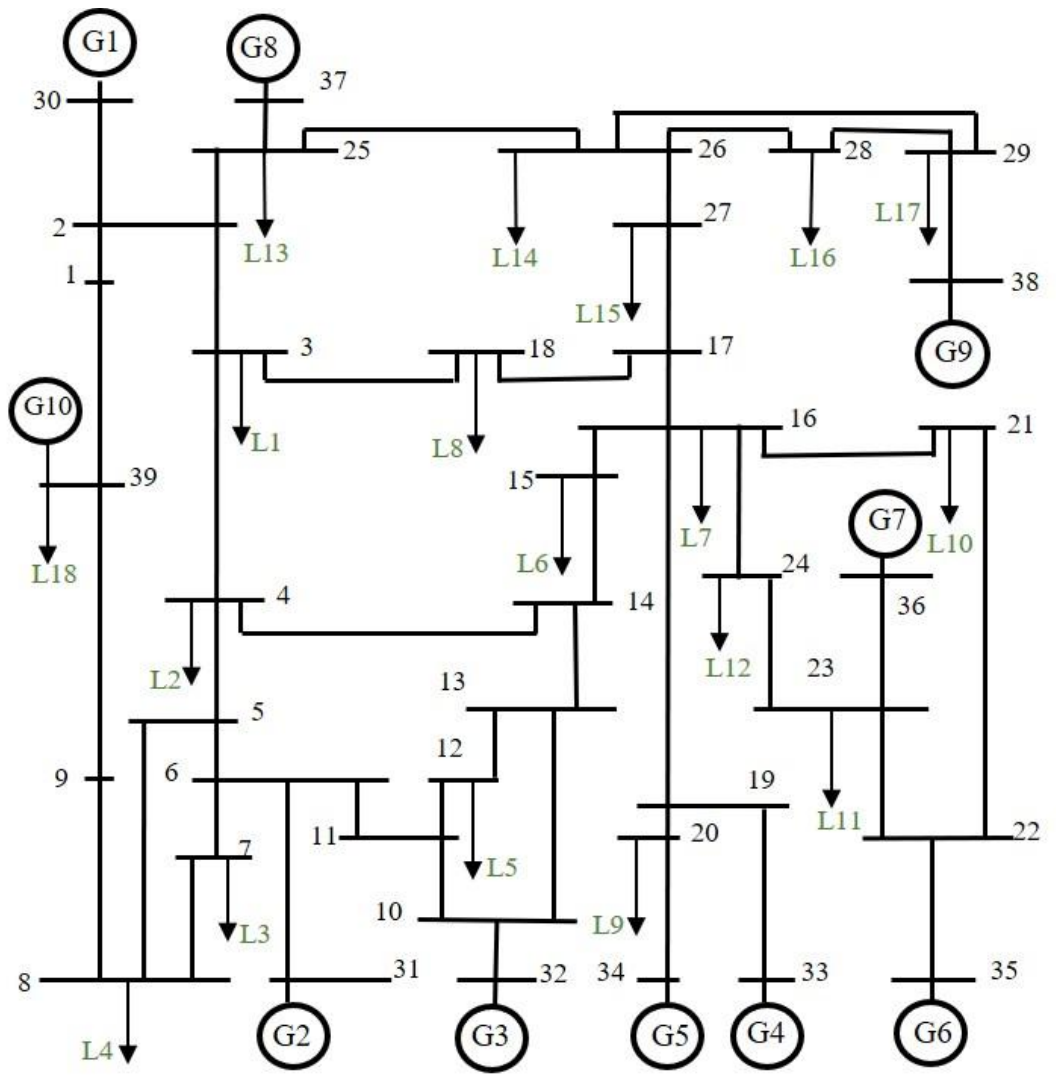}
    \caption{Topology of the IEEE 39-bus network.}
    \label{fig:IEEE39}
\end{figure}

\begin{table}[H]
\centering
\caption{Generator parameters.}
\label{tab:generator_parameters}
\renewcommand{\arraystretch}{1.2}
\setlength{\tabcolsep}{6pt}
\small
\begin{tabular}{ccccccc}
\hline
\makecell[c]{Generator\\ID} &
\makecell[c]{Bus} &
\makecell[c]{$a$ (MC-curve\\parameter)} &
\makecell[c]{$b$ (MC-curve\\parameter)} &
\makecell[c]{Maximum generation\\output (MW)} &
\makecell[c]{Maximum ramp\\rate\\(MW/h)} &
\makecell[c]{Start-up\\cost\\(\$)} \\
\hline
G1  & 30 & 14.5 & 72.4 & 1040 & 1040 & 724   \\
G2  & 31 & 16.7 & 83.4 & 646  & 646  & 487.6 \\
G3  & 32 & 16.1 & 80.6 & 725  & 725  & 535   \\
G4  & 33 & 16.6 & 83.0 & 652  & 652  & 491.2 \\
G5  & 34 & 17.9 & 89.7 & 508  & 508  & 404.8 \\
G6  & 35 & 16.4 & 82.1 & 687  & 687  & 512.2 \\
G7  & 36 & 17.2 & 86.2 & 580  & 580  & 448   \\
G8  & 37 & 17.4 & 87.1 & 564  & 564  & 438.4 \\
G9  & 38 & 15.3 & 76.6 & 865  & 865  & 619   \\
G10 & 39 & 14.2 & 71.2 & 1100 & 1100 & 760   \\
\hline
\end{tabular}
\end{table}

\noindent
\textbf{C.2 Clearing Model: SCUC--SCED}

The market is modeled over $T=96$ time periods. In each period, every agent (generation unit) submits 10-segment stepwise bids, and the market imposes a uniform price cap of $P^{\max}=150\ \$/\mathrm{MWh}$. In each episode, the environment jointly solves a network-constrained two-stage clearing process over all 96 periods: first, security-constrained unit commitment (SCUC), and then, given the unit on/off statuses, security-constrained economic dispatch (SCED). The core decision variables include the unit commitment status $u_{i,t}$, start-up variable $y_{i,t}$, segment-level generation output $g_{i,b,t}$, total generation output $g_{i,t}$, and line-relaxation variables $s_{l,t}^{+}$ and $s_{l,t}^{-}$.

The objective function of the SCUC stage can be written as:
\begin{equation}
\min \sum_{t=1}^{T}\left[
\sum_{i=1}^{N}\bigl(C_i^{\mathrm{start}} y_{i,t}+C_i^{\mathrm{fix}} u_{i,t}\bigr)
+\sum_{i=1}^{N}\sum_{b=1}^{10} p_{i,b,t} g_{i,b,t}
+\sum_{l=1}^{L} M_l\bigl(s_{l,t}^{+}+s_{l,t}^{-}\bigr)
\right]
\label{eq:scuc_objective}
\end{equation}

where $N=10$ and $L=46$, and $M_l$ denotes the penalty coefficient for line-relaxation slack. The associated constraints include:

i. Power-balance constraints
\begin{equation}
g_{i,t}=\sum_{b=1}^{10} g_{i,b,t},
\qquad
\sum_{i=1}^{N} g_{i,t}\geq \sum_{m=1}^{18} d_{m,t}
\label{eq:scuc_balance}
\end{equation}

ii. Feasibility constraints on segment-level generation output
\begin{equation}
0\leq g_{i,b,t}\leq \bar{g}_{i,b,t}\,u_{i,t}
\label{eq:scuc_segment}
\end{equation}

where $\bar{g}_{i,b,t}$ is obtained from the differences between adjacent segment boundaries of the stepwise bids.

iii. Network power-flow constraints

Let $A$ denote the bus--generator incidence matrix, $B$ the bus--load incidence matrix, and $\mathrm{SF}$ the shift-factor matrix computed from line reactances and the bus--branch incidence matrix. Then, the line power flow is given by

\begin{equation}
f_t=\mathrm{SF}\bigl(A g_t-B d_t\bigr)
\label{eq:scuc_flow}
\end{equation}

subject to
\begin{equation}
-\mathbf{F}_{t}^{\max}-s_t^{-}\leq f_t\leq \mathbf{F}_{t}^{\max}+s_t^{+},
\qquad
s_t^{+},s_t^{-}\geq 0
\label{eq:scuc_flow_limit}
\end{equation}

iv. Unit ramping constraints
\begin{equation}
g_{i,t}-g_{i,t-1}\leq RR_i
\label{eq:scuc_ramp}
\end{equation}

Subsequently, in the SCED stage, with $u_{i,t}$ fixed (i.e., with the unit commitment status fixed), the start-up cost term is removed, and only the segment-level generation output and network constraints are re-optimized so as to obtain the economic dispatch results and locational marginal prices (LMPs).

\noindent
\textbf{C.3 Calculation of Locational Marginal Prices}

In SCED, the code computes the locational marginal prices (LMPs) from the dual variables associated with the power-balance constraint and the transmission-line constraints. Let $\lambda_t^{E}$ denote the dual variable of the system energy-balance constraint, and let $\mu_t^{+}$ and $\mu_t^{-}$ denote the dual variables associated with the upper and lower transmission-line limits, respectively. Then, the nodal congestion component and the nodal price can be written as

\begin{equation}
\lambda_t^{C}=-\mathrm{SF}^{\top}\bigl(\mu_t^{+}-\mu_t^{-}\bigr),
\qquad
\lambda_t=\mathbf{1}\,\lambda_t^{E}+\lambda_t^{C}
\label{eq:lmp_decomposition}
\end{equation}

where $\lambda_t\in\mathbb{R}^{39}$ is the LMP vector for the 39 buses.

\end{document}